\documentclass[journal]{IEEEtran}
\usepackage{cite}
\ifCLASSINFOpdf
  \usepackage[pdftex]{graphicx}
  \else
\fi
\usepackage[cmex10]{amsmath}

\usepackage[font=footnotesize]{subfig}
\usepackage{fixltx2e}
\usepackage{url}

\usepackage[section]{placeins}

\begin{document}
\title{Evolution of Sustained Foraging\\ in 3D Environments With Physics}

\author{Nicolas~Chaumont
        	and~Christoph~Adami
\thanks{N. Chaumont is with the Keck Graduate Institute of Applied Life Sciences,
	Claremont, CA, 91711 USA e-mail: chaumont@msu.edu}
\thanks{C. Adami is with the Department of Microbiology and Molecular Genetics and
 		the Department of Physics and Astronomy at 
		Michigan State University,
		East Lansing, MI 48824 e-mail: adami@msu.edu}
\thanks{Manuscript received December 19, 2011; revised December 19, 2011.}}

%

\markboth{IEEE Transactions on Evolutionary Computation}%
{Shell \MakeLowercase{\textit{et al.}}: Bare Demo of IEEEtran.cls for Journals}

\maketitle

\begin{abstract}
Artificially evolving foraging behavior in simulated legged animals has proved to be a notoriously difficult task. Here, we co-evolve the morphology and controller for virtual organisms in a three-dimensional physically realistic environment to produce goal-directed legged locomotion.  We show that following and reaching multiple food sources can evolve de novo, by evaluating each organism on multiple food sources placed on a basic pattern that is gradually randomized across generations. We devised a strategy of evolutionary ``staging", where the best organism from a set of evolutionary experiments using a particular fitness function is used to seed a new set, with a  fitness function that is progressively altered to better challenge organisms as evolution improves them. We find that an organism's efficiency at reaching the first food source does not predict its ability at finding subsequent ones because foraging efficiency crucially depends on the position of the last food source reached, an effect illustrated by ``foraging maps" that capture the organism's controller state, body position, and orientation. Our best evolved foragers are able to reach multiple food sources over 90\% of the time on average, a behavior that is key to any biologically realistic simulation where a self-sustaining population has to survive by collecting food sources in three-dimensional, physical environments.

\end{abstract}


%
\IEEEpeerreviewmaketitle

\section{Introduction}
%
%
%
%
\IEEEPARstart{F}{oraging} is essential to the survival of countless animal species, and involves a variety of 
	behaviors ranging from basic chemotaxis in bacteria~\cite{SzurmantOrdal2004} and \emph{C. elegans}~\cite{ward1973chemotaxis} 
	to elaborate cooperative behaviors, for example in group hunting used by predators such as lions \cite{stander1992cooperative}
	or bottlenose dolphins~\cite{gazda2005division,sargeant2005specialization}. 
	Despite field-specific nuances, foraging involves three basic skills: Sensing and locating a target, approaching it 
	so that it is within reach, and finally handling it in some manner. 
	In biology, foraging implies the search for a resource, usually food. Then, either it is consumed
	on site (such as nectar for hummingbirds)  or it is brought back to the nest (e.g., in bees), in which case foraging also involves homing.
	The inherent link between the forager and the resource foraged stems from the necessity to develop 
	some type of morphological, physiological or behavioral apparatus for finding, reaching and processing the resource.
	This forager-resource interaction is at the origin of a variety of fundamental 
	phenomena such as mutual forager-resource co-evolution~\cite{pauw2009flies,muchhala2009going}, 
	 predator-prey co-evolutionary arms race~\cite{abrams2000evolution,toju2006imbalance},
	or the emergence of migratory, nomadic, and sedentary behaviors 
	\cite{dingle2007migration,maher2000review,mueller2008search}. Foraging can also give important insights
	into ecosystem stability and food web dynamics\cite{beckerman1997experimental,kondoh2003foraging,thierry2011consequences}.
	
	The study of foraging strategies has lead to many advances, and profoundly influenced engineering and computer science. 
	The method of stochastic optimization is a prime example of this influence, with the development of 
	well-established search algorithms such as ant colony optimization (ACO)~\cite{dorigo1999ant}, as well as particle swarm optimization (PSO)~\cite{kennedy1995particle}; 
	more recent search heuristics based on foraging concepts are the  bacterial foraging optimization algorithm (BFOA)~\cite{passino2002biomimicry},
	and artificial bee colony (ABC)\cite{karaboga2005idea}.
	Information foraging theory (IFT), which is a direct application of optimal foraging theory, 
	has deeply altered human-computer interaction research~\cite{pirolli2007information, fu2007snif}, 
	and has recently been applied to software engineering~\cite{scaffidiunifying}.
	Coordination displayed by foraging social insects inspire new heuristics in robotics for navigation~\cite{walker1993bee}, 
	homing~\cite{gaussiera2000visual}, and division of labor~\cite{Labella2006Division}.
	
	In biology, despite contributing to the mechanistic explanation of foraging behavior 
	at the genetic~\cite{pereira1993mutations, ingram2005task, page2000genetic, sokolowski1997evolution}, 
	molecular~\cite{hills2006animal}, and neural levels~\cite{gray2005circuit, grillner2008neural},
	the study of foraging and its evolution has not sufficiently improved our understanding of this phenomenon to the point where it is reproducible in artificial legged animals. 
	One reason for this failure may be that goal-directed behavior in higher animals has taken a very long evolutionary route,
	by gradually evolving from ancient instances in organisms that can be traced back to the first prokaryotes~\cite{hills2006animal}.
	Therefore, studying the evolution of foraging is mostly concerned with comparative genomics, physiology, and neuroscience, 
	and provides few if any clues about the necessary evolutionary pressures, as well as behavioral milestones, that 
	enable the evolution of dependable, sustained foraging behavior. 
	
	For this reason, the reproduction of foraging in an artificial medium--for example for the purpose of studying aspects of foraging that are impossible to study in vivo--still 
	poses very serious challenges to engineers and researchers and remains a subject of very active research~\cite{ventrella1998designing,yaeger1994computational,braitenberg1986vehicles,brooks1986robust,bongard2008behavior,jain2011multi}.
	Indeed, the problem is as yet unsolved and is a classical benchmark against which newer optimization techniques are constantly tested (see, e.g.,~\cite{guo2010distributed,xu2011modeling}).
	The most obvious applications of foraging might be in robotics, because both animals and robots
	exist and interact in the same real world, are subject to the same physical laws, and because foraging 
	is the precursor for many real world applications in robotics (e.g., automatic vacuum cleaning, planetary exploration, land mine clearing, 
	bomb or hazardous material handling, rescue operations, etc.).
	Foraging in robotics research has a slightly different meaning than in biology, and refers to the task of reaching 
	one ore more objects, collecting them, and taking them to one or several locations. Homing is almost always implied.
	Because of technical limitations, the idea of using the collected object as a resource to provide additional 
	autonomy to the robot, such as a usable spare part for self repair or connecting an additional battery to provide 
	additional energy autonomy (the equivalent of eating for animals), is essentially absent, except in very rare cases
	\cite{shen2009SINGO,ieropoulos2010ecobot}.
	
	An area where foraging holds an important place is the simulation of three-dimensional, physically realistic environments.
	Such environments constitute ideal prototype worlds that provide sufficient realism to test various concepts in robotics~\cite{ventrella1994explorations,LipsonPollack2000,bongardZykovLipson2006,Shen2008deformable-wheel}.
 Simulationis in 3D physically realistic environments are much easier to set up compared to their hardware counterpart, they can be executed many times faster than real time, and allow the same experiment to be reproduced perfectly. 
	Physically realistic simulated environments are also extensively used outside the field of robotics, such as the simulation of physical phenomena in computer graphics 
	\cite{chentanez2011real,baraff1998large,faloutsos2001virtual,zheng2010rigid,chadwick2011animating}, 
	and gaming~\cite{schrum2011},  to name but a few.
	In biology, such physically realistic environments can be used to study the evolution and usage of physical and behavioral traits that are specifically adapted to overcome the constraints dictated by the simulated laws of physics, and can help better understand how similar traits evolved and are used by real organisms~\cite{miller1988motion,shim2003generating,iwadate2011evolving,nakamura2011acquisition}.
	
	The path to the evolution of foraging in 3D legged virtual organisms is lit by several beacons, such as the evolution of forward locomotion~\cite{Sims1994b, KomosinskiUlatowski1999,Miconi2008,LassabeLugaDuthen2006,ChaumontEgliAdami2006,AuerbachBongard2010},
	competing for a single food source \cite{Sims1994b}, and the evolution of light following \cite{Sims1994a,PilatJacob2010}. 
	However, the last step, namely goal-directed movement to multiple arbitrary locations has not been achieved in this medium (but it is possible to achieve it in non-physical environments, see for example~\cite{Elsberryetal2010}).  
	Such an advance would allow a whole population of foraging organisms inhabiting a 3D physical word to autonomously feed, eventually reproduce, and evolve to produce a fully sustainable population that could even radiate into distinct species and give rise to ecosystems. This kind of simulation would enable the study of a variety of aspects related to evolutionary population dynamics, such as modes of speciation, mate selection, group formation, communication and ecosystem stability, within a framework that spans several levels of organization ranging from the response of a single neuron all the way up to the dynamics of an entire ecosystem.
	
	In this work, we report the first known instance of evolved 3D legged organisms subject to the laws of physics 
	that are able to forage food items randomly placed many body lengths away from them, and reliably repeat this performance multiple times. We expect that such foragers are able to survive autonomously in an open-ended environment where they can reproduce after gathering enough food sources, and found a sustainable population.

\section{Methods}
	\subsection{EVO} 
		The work presented here has been carried out with an updated version of the EVO platform 
		used in \cite{ChaumontEgliAdami2007} and \cite{ChaumontEgliAdami2006}. 
		Even though it remains functionally very similar to the original version, there are several design improvements.
		The software is organized around a plugin architecture where plugins implement one or more functionalities defined 
		by a public API. In a typical experiment, the user creates objects using EVO's abstract factory, which in turn
		looks for the plugins that are able to create objects of the requested types and names. By default, the user can access 
		all the pre-defined APIs in EVO, which is usually enough in most cases. Additionally, the plugins can also specify
		APIs of their own that other plugins can take advantage of, which allows EVO to be extended arbitrarily.
		
		By taking advantage of this architecture, the platform is able to select the best objects 
		when several candidates could carry out the same function, through a variant of the chain-of-responsibility pattern. 
		This feature allows EVO to automatically select appropriate plugins 
		to load an unknown organism from the hard drive, interpret its genetic encoding to produce a functioning agent, and mutate 
		and recombine it without having to recompile. Multiple organisms with different morphologies and controllers can then
		be loaded, simulated, and genetically altered simultaneously within the same competitive environment.
		The following sections cover the differences with the initial implementation in Refs.~\cite{ChaumontEgliAdami2006,ChaumontEgliAdami2007}.
		
	\subsection{Virtual organisms}
		The virtual organisms used in this work are very similar to the walkers used in \cite{ChaumontEgliAdami2006} and \cite{ChaumontEgliAdami2007}. 
		The main differences lie in the genome encoding and the neural controller. 
		
		This version of EVO provides a default grammar and a set of 13 keywords 
		to describe the genome structure, and the rules to initialize and mutate it. 
		The genome description is specified in a script-like fashion (see Supplementary Material for a description of the script~\footnote{An electronic supplement for this manuscript is available at \\
		http://tinyurl.com/autonomous-foragers-supplement}) and stored in a file.
		Once read, the description is interpreted and stored in data structures that are used by the plugins in charge of initializing 
		and mutating the organism's genome. Without this script, the user would have to code by hand, compile, and debug the functions that initialize 
		and mutate each new genetic structure, and go back to the code each time the structure is modified, which is laborious. 
		These modifications do not involve any coding or re-compilation in the current version of EVO, and the genetic description can be written much faster than if coded, 
		typically in a few minutes.

		The controller used in this work has more neural types compared to the 2006-2007 version, implementing all those used by Sims~\cite{Sims1994a,Sims1994b}. The types are defined by the manner in which they process inputs: Sum, Product, Divide, SumThreshold, GreaterThan, 
		SignOf, Min, Max, Abs, If, Interpolate, Sin, Cos, Atan, Log, Exp, Sigmoid, Integrate, Differentiate, Smooth, Memory, Wave, Saw, and constant. 
		In addition to the contact sensors, each joint has an angular proprioceptive sensor. Furthermore, the root block---from which the rest of the morphology
		is created---has two sensors that return the angle and the distance to the closest food source. 
		The angle is measured between the organism's forward vector and the vector that originates from the organism's root block center of gravity, 
		and points to the closest food source.
	
	\subsection{Evolutionary algorithm} 
		The evolutionary algorithm used in this work is a steady-state genetic algorithm (SSGA)~\cite{WhitleyKauth1988,Syswerda1989}
		that starts by evaluating an initial population of 200 organisms.
		The evaluation phase typically consists of simulating each individual in the population 
		with the same set of targets and calculating the geometric mean of the fitness obtained for each target.
		Since the individual and its environment are destroyed and recreated anew for each simulation, 
		the organism retains no history about its past experience with previous targets.
		
		Once every organism is evaluated, in the selection phase 20\% of the population survives,  
		and is used to repopulate the vacant 80\%. Three selection strategies can be used to choose a survivor: 
		elite, roulette, and tournament. In elite selection, the organisms with the highest fitness in the population are guaranteed to survive, while in roulette selection the organisms with higher fitness have a higher probability to leave offspring. In tournament selection, the best of a group of organisms (with given tournament size) is selected to populate the next generation, with as many repeats of the tournament as necessary to fill up the population.
		Each method is responsible for selecting a fraction of the survivors so that all three methods 
		can be used simultaneously within the same selection round. 
		Moreover, to keep the maximum possible diversity, organisms selected with one method cannot be selected by another, but could be picked multiple times 
		within the same selection strategy, as is the case for roulette and tournament. 
		The order in which selection strategies are used is then important, as the last method used cannot choose
		organisms picked by the previous methods, and has therefore less individuals to choose from.
		Their default order is implemented as follows:  Elite is used first, then roulette, and finally tournament. 
		This order, however, can be overridden by an arbitrary user-defined sequence.
		
		The reproduction phase is very similar to that of Sims \cite{Sims1994a,Sims1994b}: 
		An empty slot is filled with an organism generated from the survivors.
		The new individual is either a clone with 30\% probability, or a recombination between two survivors. 
		In either case, the new organism undergoes a 1\% per-site mutation rate, 
		which corresponds to about 1 to 10 mutation events per genome.
		Once reproduction is completed, the algorithm goes through a new evaluation phase 
		and repeats the same sequence for typically 40 to 50 generations. Such a short evolutionary time is sufficient to witness the emergence of promising foraging behaviors without spending computing power on poor organisms.

\begin{figure}[!ht]
		\begin{center}
			\includegraphics[width=0.4\textwidth]{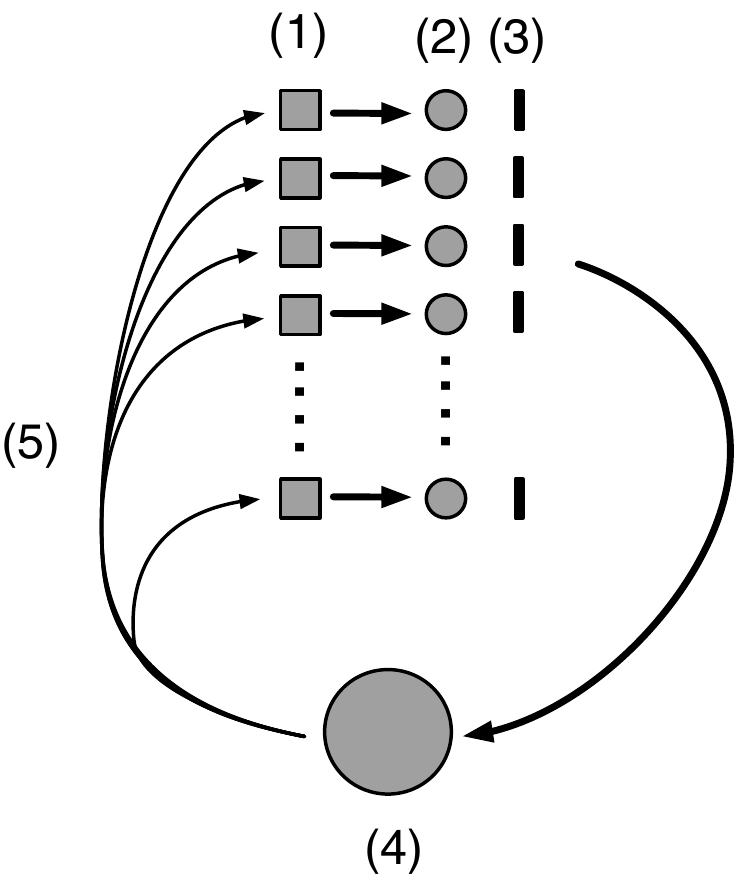}
		\end{center}
		\caption{
		{\bf Transfer strategy.} 
		(1): At any stage of the evolutionary process, a large number of simulations are executed in parallel (gray rectangles).
		(2): Each simulation (or repeat) yields an evolved organism (gray circles).
		(3): All the evolved organisms are visually inspected for desired foraging properties, 
		where only a single organism across each repeat passes the inspection (illustrated by the dashed line).
		The best organism among all the evolved individuals from these runs is called a \emph{key organism} (4),
		and is used to seed the next set of runs (5). The transfers are repeated five times, 
		as is illustrated by the line of descent in Fig.~\ref{fig:LOD}.
		}
		\label{fig:Transfer_strategy}
		\end{figure}
		Unlike the emergence of forward locomotion that is systematic in EVO~\cite{ChaumontEgliAdami2007}, foraging builds upon skills that are much more rare, and thus less likely to evolve. 
		For instance, while forward locomotion emerges robustly within 20 generations, 
		only 3\% of the simulations (out of 400 repeats) successfully evolved organisms that took at least two different 
		trajectories when exposed to different food sources after 40 generations. 
		This skill, namely the ability to exhibit a behavior that is conditional on food source placement, is a necessary milestone before reliable foraging can emerge. Yet, over 30 variations of the same experiment run for 40 generations with the present set-up would have to be repeated on average before it appears in only one of them, on average.
		For this reason, each simulation configuration is run in sets of 100 to 480 repeats with the same parameters save the random number seed. Each repeat produces an evolved organism that we inspect visually and whose performance is compared against the organisms from the other repeats (see Fig.~\ref{fig:Transfer_strategy} for an overview of the transfer strategy) 
		
\begin{figure*}[!ht]
		\begin{center}
			\includegraphics[width=0.6\textwidth]{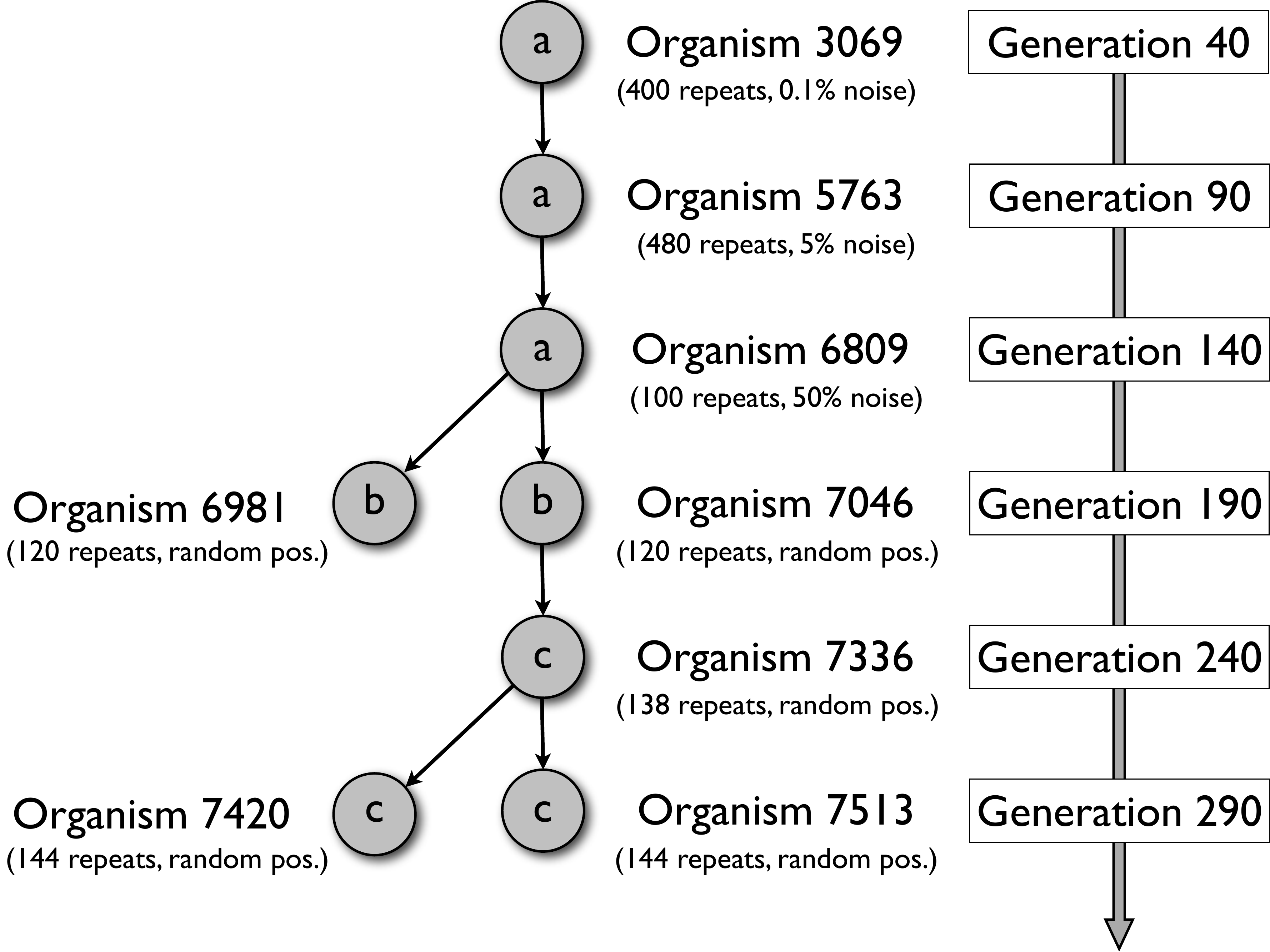}
		\end{center}
		\caption{
		{\bf Line of descent of the key organisms that led to the best foragers.} 
		Each individual is a key organism, except organism 6981, which is shown for the sole purpose of illustrating its position in the line of descent. Key organisms are used as seed for the next stage.
		The numbers in parenthesis below each organism indicates how many repeats of the same experiment were executed in parallel in the stage that gave rise to it,
		as well as the amount of noise applied to the food source positions. 
		The letters in each gray circle indicate which type of fitness function was used to evolve the corresponding key organism: 
		fitness (a) is computed with Eq.~(\ref{eq:Wrepeat}), fitness (b) with Eq.~(\ref{eq:Wrepeat2}), 
		and fitness (c) is the total number of food sources reached by the organism.
		The cumulative number of generations needed to obtain a specific organism is shown on the right.
		The uneven number of generations between key organisms 
		  and the variation of repeat sizes are the result of optimally balancing 
		the computational load at the time of submitting new jobs given the maximum limit of 144 simultaneously available CPU cores on the HPC cluster.
		}
		\label{fig:LOD}
		\end{figure*}

		The best evolved organism from a set of repeats is called  a \emph{key organism} because it exhibits key behavioral qualities that are desirable for foraging, such as locomotion in the early stages of evolution, and the flexibility to steer towards multiple directions in later generations.
		For example, the first key organism (organism 3069) evolved after 40 generations from a population of organisms seeded with random initial genotypes, and
		was selected among the fittest from 400 repeats (from run 2986 to run 3385, see Fig.~\ref{fig:LOD}). 
		Each repeat is numbered according to the order in which it is executed, 
		and each evolved organism is numbered after the evolutionary experiment that produced it.
		The first key organism is then used for the next set of 480 repeats, 
		where each population in the repeats is composed of exact clones of organism 3069.
		This procedure is repeated five times (with different numbers of repeats, and different appropriate fitness landscapes) to yield the final foragers (see Fig.~\ref{fig:LOD}).

	\subsection{Placement of food sources}
		A critical aspect of autonomous foraging is the organism's ability to use the relevant sensory information to approach and reach nearby targets. 
		To evolve such a trait, the organism has to be exposed to a large number of different target positions, which is why in this work we vary 
		those locations within and across generations.
		When evaluating the fitness of a single organism, it is impractical to uniformly sample enough random target positions 
		to recreate a representative approximation of the task of foraging. Indeed, with the current setup, we would need to test each organism on at least 20 to 30
		different targets to cover the available space, which corresponds to several weeks of simulation for a single experiment. 
		However, in the early stages of evolution, where the organisms have a very limited ability to approach 
		food sources, their fitness is overwhelmingly dictated by the targets' positions: the fittest individuals are therefore those for which the greatest 
		number of targets (usually one) happen to be at the right place. Therefore, the fittest are, at that stage, not the most efficient foragers but instead they are the luckiest ones.
		As the foragers become increasingly capable, their dependence on the target positions dwindles to the point where the fittest 
		individuals are almost always the most efficient ones.
		
		We use an approach inspired by this observation to reduce the stochastic effect of random food source positions while keeping 
		the simulation time to a minimum (within a day to a week per experiment): We first find a target placement that yields the fastest emergence 
		of target-following behaviors within a fixed amount of time,
		then we use this placement in conjunction with an increasing amount of noise to allow the organisms to reach food sources at arbitrary positions. 
		The optimal placement requires only four food source locations compared to 20 or 30 as discussed earlier.
		A significant drawback to this approach is the fitness sensitivity to stochastic fluctuations on the food source placements.
		Those random fluctuations can be mitigated by evolving the initial population in an almost noise-free environment, and gradually increasing the noise for each subsequent evolutionary stage once the best organism's performance is reliable given the previous amount of noise.
		
\begin{figure}[!ht]
\includegraphics[width=0.5\textwidth]{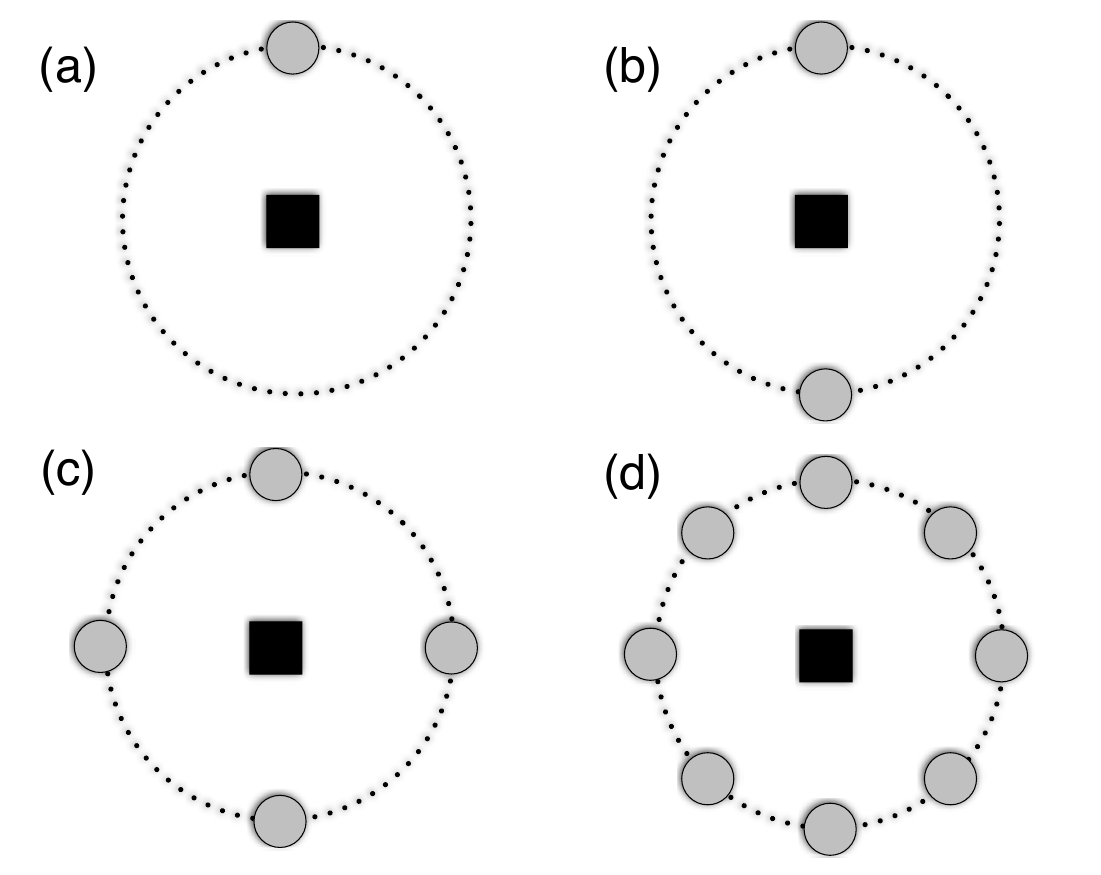}
\caption{
		{\bf Food source placement before the application of noise.} 
		Depending on the number of evaluation steps used to evaluate an organism, the food sources \emph{across} evaluation steps are uniformly spaced at constant angle intervals on the perimeter of a circle, starting with the front position first. Within a conditional movement adaptation step, all three food sources have strictly the same coordinates before noise is applied. Several different conditional movement selection scenarios can be used, which have different tradeoffs between noise adaptation, conditional movement adaptation, and computation time. (a): A single conditional movement adaptation step, used to select for adaptation to noise only. (b): Two conditional movement adaptation steps, selecting for conditional movement forwards or backwards. (c): Food source placement when selecting for conditional movement in the four cardinal directions. (d): Selection scenario using eight primary locations for conditional movement selection. Using such a placement can obviate the use of large random deviations (thus saving a transfer step described in Fig.~\ref{fig:LOD}) as they would be covered by the primary directions, but at the expense of testing many locations for every single organisms.
		}
\label{fig:evaluation_steps}
\end{figure}

\begin{figure}[!ht]
\includegraphics[width=0.5\textwidth]{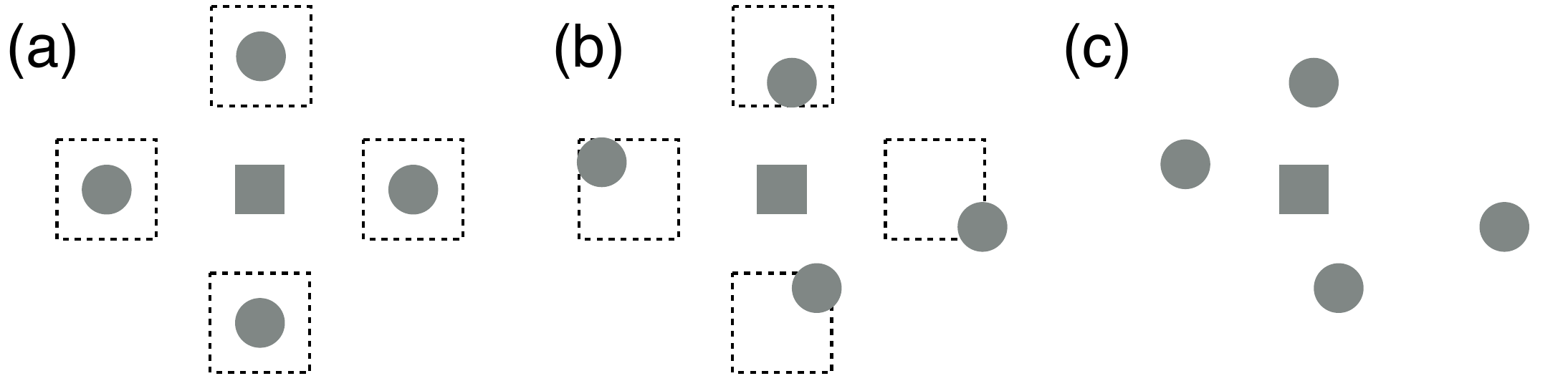}
\caption{
		{\bf Application of noise to the food source locations.}
		The maximum amount of perturbation applied to the food sources is a function of the distance between the target (solid circle) 
		and the organism (solid square).
		The perturbation is uniformly added to both the $x$ and $y$ coordinates of the food sources, which is equivalent to drawing squares
		around them (a) and uniformly choosing a position within the boundary of those squares (b). (c): Resulting positions after noise is applied.
		}
		\label{fig:noise_application}
		\end{figure}

The algorithm for the placement of food sources has two components. First, we select for the ability to move conditionally on food source location, by positioning the source 10 meters away from the organism (measured between the organism's root block and the food source's centers of gravity), at a specific angle. For example, by choosing two locations (angles 0 degrees and 180 degrees), we select to move forward when the food source is located in front, but move backwards (or turn around and then move forwards) when the source is in the back (see Fig.~\ref{fig:evaluation_steps}b). We can also use the four cardinal directions to select for conditional movement (Fig.~\ref{fig:evaluation_steps}c) or more  (Fig.~\ref{fig:evaluation_steps}d). Within each 
conditional movement adaptation step, we perform a {\em noise adaptation step}, where we randomize the source placement by applying noise along the horizontal plane, by an amount that is proportional to the distance used in the first step (Fig.~\ref{fig:noise_application}).  This step is important in the early transfer stages of evolution (defined in Fig.~\ref{fig:LOD}) because very little noise is used in those: in that case the food source positions carry a strong and consistent signal of directionality. 
As the foragers become more efficient, the noise component is increased up to the point where the noise adaptation step is meaningless and is skipped altogether. For the noise adaptation step, we repeat each primary direction (defined by the angle) three times, so that an organism can reach up to three food sources in sequence (with positions that differ only because noise is applied). Note, however, that the internal state of the organism is not reset when the experiment is repeated; instead we are really asking the organism to reach up to three food sources {\em in sequence}. We chose these two different evaluation steps because adaptation to noise and adaptation to drastically different locations appear to be different skills: it is possible, for example, to evolve to conditionally react to sources right in front and to the back, and still be incapable to tolerate even the smallest deviations from these two positions. At the same time, it is possible to tolerate noise in one direction, but not be capable to conditionally react to arbitrary directions.
		
Each fitness evaluation in the noise adaption step yields a fitness value $W$ for reaching up to three food sources in sequence [see Eq.~(\ref{eq:Wa2}) below]. This fitness is then combined with the rest of the other fitness values (obtained in the conditional movement steps) in a geometric average to produce a composite fitness $\overline{W}$ [Eq.~(\ref{eq:Wbar})]. The order of an evaluation step is going to determine the angle at which a food source is placed: 
		In the first conditional movement step, all the food sources in the sequence are in the same location, 10 meters in front of the organism (possibly with noise). In each subsequent conditional movement step, the food sources are placed at an angle that is determined by the order of the evaluation. 
		All the food sources within a sequence initially overlap each other (before noise is applied). When the organism reaches a target, the next target coordinate is added to the current organism's position.  
		
For instance, let us consider an organism evaluated across four evaluation steps (Fig. \ref{fig:evaluation_steps}c), so that each angle interval is $(2 \pi) / 4 = \pi/2$ radians. We test the individual with a maximum of 3 food sources, and would create the first one in front, at a 0 degree angle (possibly with noise). If the forager can reach the first food source on time, then the second target position is added to the organism's current position and appears at a 0 degree angle (plus noise) as well, as would the third target if the second one were reached. After the first conditional movement step terminates, the second such step begins with a food source placed at an angle of  $\pi/2$ radians, i.e., on the left. Each new food source (up to three) within this conditional movement step will appear at the same angle, 10 meters away from the organism. The last two steps will create food sources at angles of $\pi$ radians and $3 \pi / 4$ radians, which correspond to food sources in the back and on the right, respectively. At any point in time, an individual sees only one food source.
		
When noise is applied to the food source location, each target position within a sequence is uniformly randomized by adding noise $\nu$ proportional to a percentage $p$ of the distance to the food source $\delta$, so that $\nu = \delta \cdot U(p, -p)$, where $U(x, y)$ is the uniform distribution over a 2D plane spanned by the $x$ and $y$ axes. For instance, a noise level of 5\% applied to a food source placed 10 meters away from the individual corresponds to two uniformly distributed perturbations of at most $\pm$50 cm added to the target's $x$ and $y$ coordinates. This is equivalent to drawing squares of $1 {\rm m}^2$ centered on each food source, and choosing a position uniformly within the perimeter of those squares (Fig.~\ref{fig:noise_application}).


	\subsection{Simulation}
		Before evaluation, each individual has to successfully pass a {\em validity test} that discards those that are either prone to create numerical
		instabilities within the physics simulator, or those that have no chance of moving or sensing. To determine the latter, an organism is removed from the simulation 
		if it has only one block, if its neural network is disconnected from the motors, sensors, or both, or if there is any interpenetration 
		among the body parts at the very first time step. 
		This strategy significantly decreases the time spent on evaluating individuals that have no chance of succeeding~\cite{Bongard2011}. 
		The simulation creates a new empty environment, consisting of a flat surface (ground) and the first food source in the evaluation sequence. 
		Subsequently, the organism is generated from its genome description and placed above the ground as in Ref.~\cite{ChaumontEgliAdami2007}. 
		The individual's proximity and angular sensors are initialized to target the food source. 
		The physics in the virtual world is simulated using the Open Dynamics Engine~\cite{ODE} in discrete time intervals of 0.02 seconds (i.e., 50 iterations per second). 
		Immediately after being created, an organism undergoes a series of activations and deactivations (described in Ref~\cite{ChaumontEgliAdami2007})
		to prevent unsuitable ``cheating behaviors" such as morphologies that manage to traverse some distance just by falling over because of the pull of gravity, 
		without using its brain at all (such cheaters were previously described by Sims in Ref.~\cite{Sims1994a}). 
		Cheaters inhibit the evolution of efficient gaits because they are much more likely to emerge by chance than individuals with cyclic, coordinated motions.
		
		As soon as the organism completes the activation sequence, a timer is initialized (at 10, 30 or 60 seconds depending on the evolution stage),
		which is the maximum amount of time given to reach the next food source before the simulation terminates.
		At each time step, the individual's root block position and the distance returned by its proximity sensor are recorded.
		When the  center of gravity of the organism's root block is within 2 meters from the food source (called the \emph{absorption radius}),
		the individual absorbs it instantaneously and the next food source in the sequence appears. 
		The proximity and angle sensors are then updated to target the new food source and the timer is reset, 
		allowing the simulation to proceed for the same amount of time as for the initial food source.
		If the organism reaches the last food source in the sequence, the food item remains in place and the simulation stops after time runs out. 
		When the simulation ends, the organism's positions and distances to the sequence of targets are used to evaluate its fitness.
	
	\subsection{Fitness Function}
		At each time step, the organism is rewarded for approaching, or penalized for moving away, from a target.
		For the current food source $s$, the fitness contribution $W_{s}$ is computed as a function of the distance 
		$\Delta d_{t} = d_{t} - d_{t-1}$ 
		travelled towards the food source between the two consecutive time steps $t$ and $t-1$. The values $d_t$ and $d_{t-1}$
		are the organism's proximity sensor readings at time $t$ and $t-1$ respectively:
		\begin{equation}
		\label{eq:Ws}
			W_s = \prod_{t=1}^{T_s-1}{(1 + \Delta d_{t})}\;,
		\end{equation}
		where $t=1\ldots T_s$ are time steps during which the food source $s$ exists in the environment. Depending on how fast
		an organism reaches the target, the variable $T_s$ is different for each food source $s = 1 \ldots N$.

		For a given number of time steps $T_s$, $W_s$ is largest when all the distances $\Delta d_t$ are equal.
		For example, in the case of two $\Delta d$ (i.e., for three time steps), the fitness can be pictured as the two-dimensional area 
		$W_s = \Delta d_1\cdot \Delta d_2$, which is maximal when $\Delta d_1 = \Delta d_2$ form a perfect square. The same reasoning
		can be applied for three $\Delta d$s that span a volume; the statement that $W_s$  is largest when all the distances $\Delta d_t$ 
		are equal can easily be proven by induction for higher dimensions. 
		This property has the advantage of favoring smooth periodic motions, because for a given number of time steps $T_s$, 
		those individuals that progress at a constant speed towards the food source are rewarded the most. 
		
		The total fitness accumulated in a simulation run for approaching targets $W_a$,
		is the product of the individual fitnesses $W_s$ for each individual food source in the sequence:
		\begin{equation}
		\label{eq:Wa}
			W_a = \prod_{s=1}^{S}{W_{s}} = \prod_{s=1}^{S}{ \left( \prod_{t=1}^{T_s-1}{(1 + \Delta d_{t})} \right) }\;.
		\end{equation}
		Unfortunately, no interesting behavior emerges with this fitness function: the best organisms move very little because 
		locomotion is not rewarded enough in the early stages of evolution. To counteract this, 
		we use an additional term that explicitly rewards locomotion using the total distance travelled 
		by an organism between the first ($p_0$) and the last ($p_f$) position recorded during simulation:
		\begin{equation}
		\label{eq:Wl}
			W_l = C \cdot \min(p_0-p_f, 1.0)\;.
		\end{equation}
		
		A scaling constant $C=100$ yields the desired results in early generations, but the term $W_l$ can become
		overwhelmingly dominant when the organisms quickly evolve to travel greater distances, 
		since it is apparently easier to develop a faster gait than to better steer towards a target.
		The result is a rapid emergence of fast movers, but poor target followers.
		To limit the influence of $W_l$, the maximum distance rewarded was capped at 1 meter.
		
		If the goal of selection is to promote target-reaching behaviors, then an organism that reaches food sources should receive a higher fitness
		if it succeeds than any other that fails. However, the fitness $W_a\times W_I$ does not explicitly reward reaching any food source per se,
		only the distance travelled towards them. 
		This implies that a forager that reaches nearby targets is penalized over those that steadily approach distant ones, without ever reaching any. 
		To correct for this, we add a term $W_r$: a bonus that is added to the fitness each time a food source is reached 
		such that organisms that get to more targets are fitter than those that attain fewer. 
		The bonus is designed such that it is about the same order of magnitude as the fitness accumulated for approaching the target. 
		Assuming the organism reaches $S$ targets in $T$ time steps where each food source is initially placed 10 meters away, 
		it will travel an average distance of $\Delta d = \frac{10\cdot S}{T}$ at each time step. 
		Substituting in \eqref{eq:Wa} and multiplying by the total number of food sources $S$, the total bonus $W_r$ for reaching
		$S$ food sources is:
		\begin{equation}
		\label{eq:Wr}
			W_r = S \cdot \left(1 + \frac{10 \cdot S}{T} \right) ^T\;.
		\end{equation}
		Since $S < 10$ and $T > 1000$, the equation can be approximated by
		\begin{equation}
			W_r \approx S \cdot e^{10 \cdot S}
		\end{equation}
		
		This bonus is added to the overall fitness:
		\begin{equation}
		\label{eq:Wrepeat}
			W = W_l \cdot \prod_{s=1}^{S}{W_{s}} + W_r\;.
		\end{equation}
		With this fitness function, the combined effect of rewarding the organism for approaching food sources with 
		the added bonus for reaching them puts a very strong evolutionary pressure for reaching the easiest target:
		As soon as an organism evolves to reach the first food source, it capitalizes on this successful direction to reach the next one 
		without developing the ability to move towards other directions. A solution is to discourage the organism to go towards
		the next food source if it already reached one in the same direction by reducing $W_s$ in Eq.~(\ref{eq:Wa}):
		\begin{equation}
		\label{eq:Wa2}
			W = \prod_{s=1}^{S}{W_{s}} = \prod_{s=1}^{S}{ \left( \prod_{t=1}^{T_s-1}{(1 + \frac{\Delta d_{t}}{2^{s}})} \right) }
		\end{equation}
		
		Each time a new food source is absorbed, the reward $\Delta d_t$ is halved. The bonus $W_r$, 
		however, remains identical, so that the organism still has a very strong incentive to reach food sources.
		Only the order of priority to reach them is changed. 
		The organism's fitness across all repeats $\overline{W}$ is the geometric average of each individual fitness $W_i$ [calculated using Eq.~(\ref{eq:Wa2})] 
		across all $R$ repeats, where $R$ is determined by the number of primary directions used in the conditional movement step described above:
		\begin{equation}
		\label{eq:Wbar}
			\overline{W} = \left( \prod_{i=1}^{R}{W_i} \right) ^{1/R}\;.
		\end{equation}

	\subsection{Measuring the organism's foraging ability}
	
		\begin{figure}[!ht]
		\includegraphics[width=0.5\textwidth]{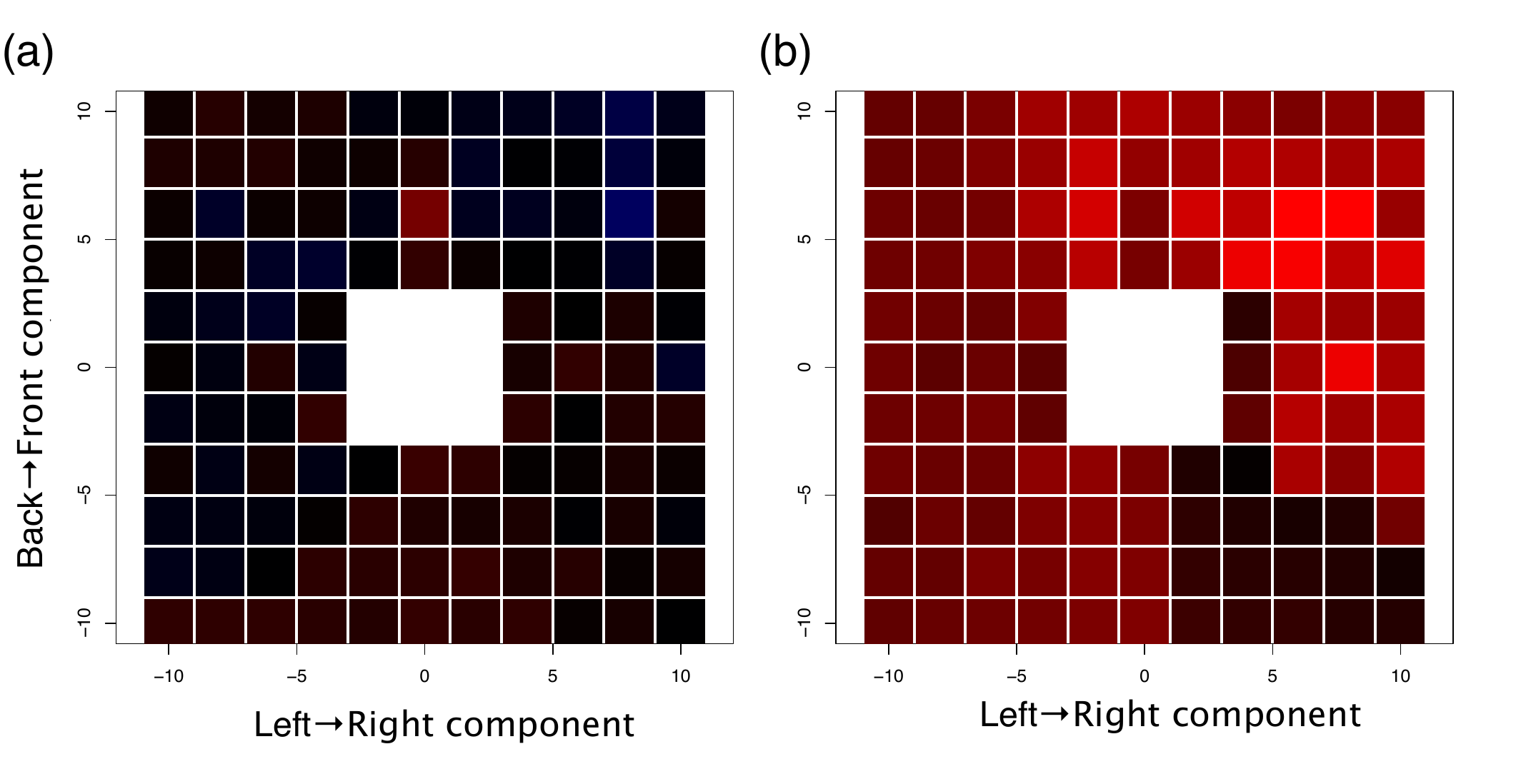}
		\caption{
		{\bf Examples of foraging maps for the first two key organisms.} (a) Organism 3069, and (b): Organism 5763. 
		Organism 3069 is the best after 40 generations from the first 400 runs. This individual was selected for its
		ability to orient itself towards the food source. There is a slim crown of lighter squares around the center of the foraging map.
		This feeble signal is the precursor for the development of steering towards more food sources in later generations. 
		The descendant from organism 3069 is on the right.
		A dark lower right octant still remains indicating that it cannot reach food sources there, 
		as if it could not interpret them (a ``zone of confusion").
		}
		\label{fig:confusing_zone}
		\end{figure}
		\hfill

		Foraging performance is assessed by placing food sources on a two-dimensional grid or map, and by testing
		the individual for each food source on that map, one at a time. Each cell on that \emph{foraging map} is colored depending on the organism's 
		average speed of approach: Lighter shades of red correspond to faster speeds \emph{toward} the food source whereas 
		lighter shades of blue encode faster speeds \emph{away from} the food source. 
		Darker shades indicate slower speeds.
		If the food source is placed close enough to the center of the grid, it would be in direct contact with the individual, 
		and be automatically absorbed even if the forager does not move at all. 
		The foraging map would display inconsistent color patterns in the center that do not reflect the organism's foraging ability at all. 
		This is why there are no food sources within 4 meters (i.e., two absorption radii) from the grid center. 
		The resulting foraging map is a snapshot of the organism's ability to forage at a given resolution.
		\begin{figure}[!ht]
			\includegraphics[width=0.5\textwidth]{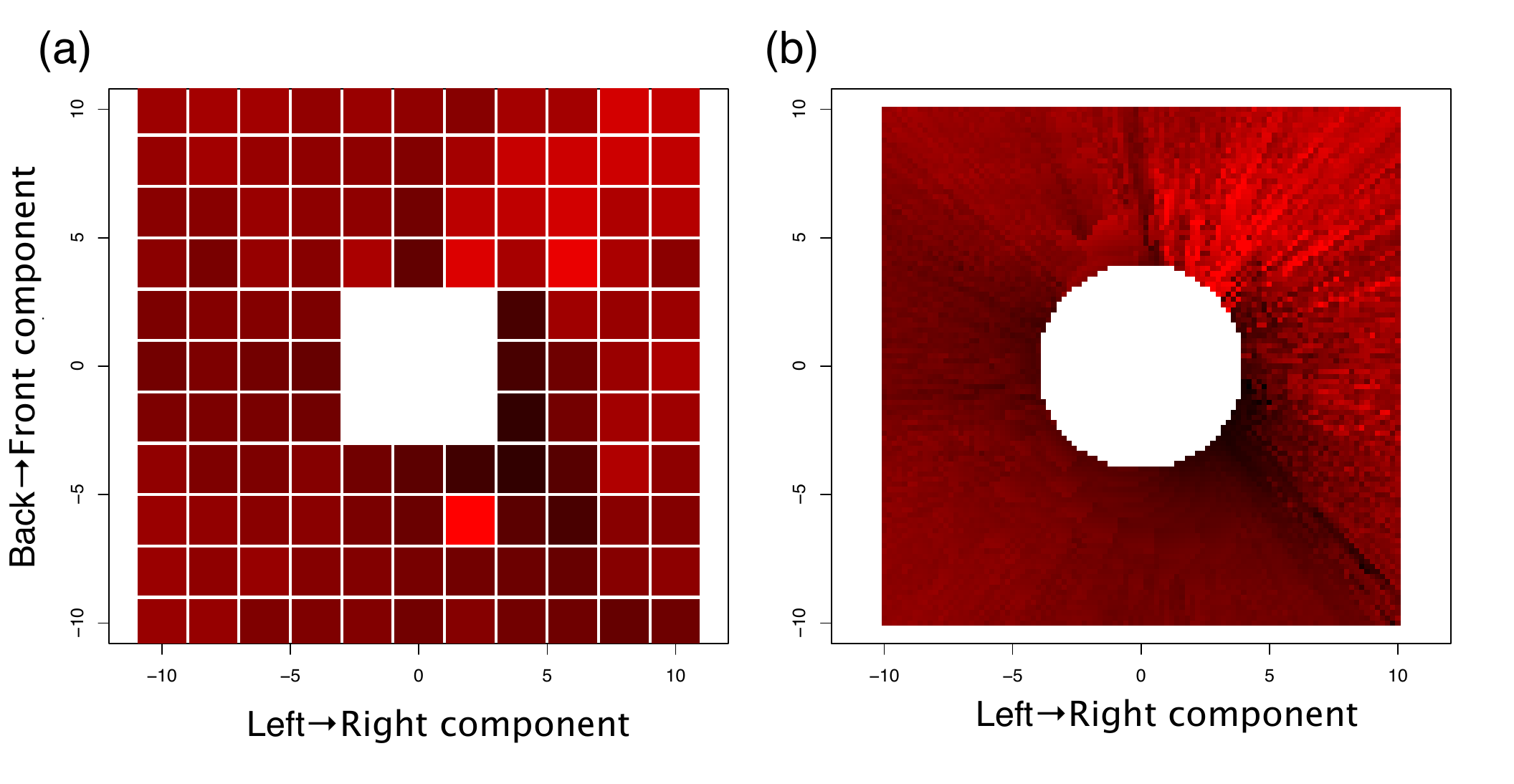}
		\caption{
		{\bf Comparison of organism 6809's foraging maps at different resolutions.} 
		Two foraging maps from a descendent of organism 3069 generated for
		(a): $11 \times 11$  and (b): $101 \times 101$ food source grids respectively. 
		Many more patterns emerge at the finer resolution, especially around the center.
		}
		\label{fig:resolution_comparison}
		\end{figure}
		
		Foraging maps in this work span a constant square surface of $20 \times 20$ meters (which is the same area used to place food items in the evolutionary runs) and carry either $11 \times 11$ (low resolution) or $101 \times 101$ (high resolution) food sources, which correspond to spacings between two adjacent food sources of 2 m and 20 cm, respectively (Fig.~\ref{fig:resolution_comparison}). 
		Because the area in the center of the foraging map is empty, an $11 \times 11$ foraging map has only 112 food sources, and not 121
		as expected in a fully populated grid.
		While the typical fitness evaluation of an organism in an evolutionary run takes about 5 seconds, the computation of a low resolution foraging map
		takes about two minutes, and about four hours for a high resolution foraging map.

		As soon as an organism reaches a food source, it is possible to draw a foraging map for all the second food sources, 
		conditional on the first one. Such a {\em conditional foraging map} provides insights about how the organism's foraging ability 
		changes after reaching specific food sources. However, the computation time approximately doubles, since for each pixel on the conditional foraging map the organism has to be simulated for two targets: the first one and all the second ones given the resolution of the map. For each conditional foraging map, the first target location is the same. Clearly, calculating {\em all} conditional foraging maps scales quadratically, so instead we present them only for selected first-target locations.

\section{Results}
	\subsection{Foraging in more than one direction}
		In order to effectively force organisms to use their sensory information, food sources have to be placed at least in opposite directions, as in Fig.~\ref{fig:evaluation_steps}b. We found that if the targets are placed in front of the organism but at different locations, each individual almost always evolves
		its own unique forward trajectory, regardless of the actual food source positions. In that way,
		by just going forward, the organisms are able to obtain (with a single unconditional behavior) some reward from all three food sources, without using their sensors at all. 
		When two targets are placed in opposite directions (Fig.~\ref{fig:evaluation_steps}b), the fittest and most effective strategies consisted in wiggling
		towards the front one, and wiggling backwards toward the target behind. 
		Both gaits are very similar and apart from the obvious progression in opposite directions, are otherwise impossible to distinguish if inspected separately.
		Very few strategies involve turning towards the food source, and those that do turn look very laborious and can never reach a 
		second target that is behind, because they are incapable of completing the turn in time. 
		Moreover, this wiggling behavior appears to be particularly inflexible: when exposed to an environment with four food sources, those individuals almost never acquire
		the ability to turn left or right, and when they do, they are always very inefficient.  
		The best results were obtained when we exposed individuals to four food sources (Fig.~\ref{fig:evaluation_steps}c). 
		We also tried exposing the individuals to eight food sources, with four additional targets on the diagonals (Fig.~\ref{fig:evaluation_steps}d), 
		but even though it yielded slightly better results for a fixed number of generations,
		such runs took twice as much time to complete: with the same amount of CPU time and the same number of replicates, 
		we had much better results by running two sets of replicates with only four food sources each, 
		where the second set is seeded with the best organisms from the first set, rather than running one set with eight food sources.
		
		If an organism has to reach one food source and the simulation terminates only when time runs out, regardless of whether it
		reaches the target or not, then distance matters: If the food source is placed far enough from the organism that it has no chance of reaching it
		within the time allowed, then evolution capitalizes on the fitness gathered from approaching a single target by optimizing the current gait to become
		ever faster, without developing the ability to go toward other directions. 
		On the other hand, if an organism can reach its first target in the early generations,
		then it cannot improve its fitness by approaching this food source any more, and is forced to adapt to seek additional food sources in other directions.
		Among the set of 400 initial runs, no organism is able to efficiently evolve a gait within 40 generations that is effective for the four directions. 
		Reasonably high fitness values (on the order of a hundred) are good indicators of efficient foraging gaits, 
		while mediocre gaits typically have fitness [measured by Eq.~(\ref{eq:Wbar})] of the order of 5. 
		An unusually fit individual (given the other member is the population), especially in the early stages of evolution, is usually a signature of a bad strategy. For example, ``jumping" organisms, that use motor impulses that violently propel them in the air, 
		can manage to randomly hop around covering greater distances than their peers in the population, 
		and happen to land on or close to a food source purely by chance. 
				
		\begin{figure*}[!ht]
		\begin{center}
		\includegraphics[width=0.8\textwidth]{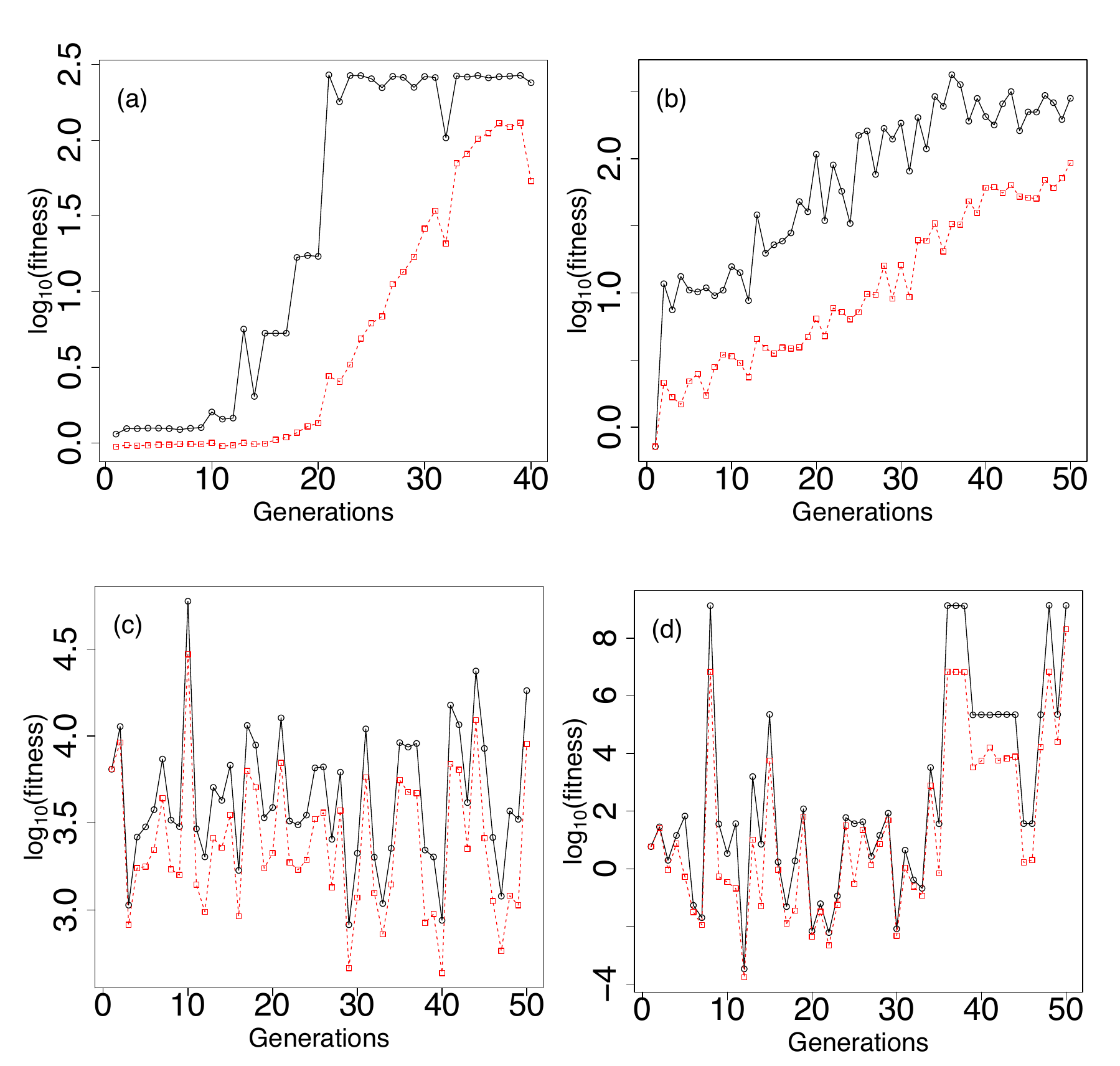}
		\end{center}
		\caption{
		{\bf Fitness profiles of evolutionary runs for various mutation rates.} Solid black lines and circles: fitness of best of population. Red dashed lines and squares: mean fitness of population.  (a): The evolutionary run that gave rise to organism 3069: 0.1\% noise, random initial population (see Fig.~\ref{fig:LOD}). 
		(b): Evolution leading to organism 5763: 5\% noise, pre-evolved population seeded by 3069. 
		(c): Evolution leading to organism 6809: 50\% noise, pre-evolved population seeded by 5763. 
		(d):  Evolution leading to organism 7046: food sources randomly placed, pre-evolved population seeded by 6809.
		Note that organism 7046 is not rewarded for the first food source.
		}
		\label{fig:organisms_fitness}
		\end{figure*}

	\subsection{Improving foraging coverage}
		\begin{figure*}[!ht]
		\begin{center}
		\includegraphics[width=0.8\textwidth]{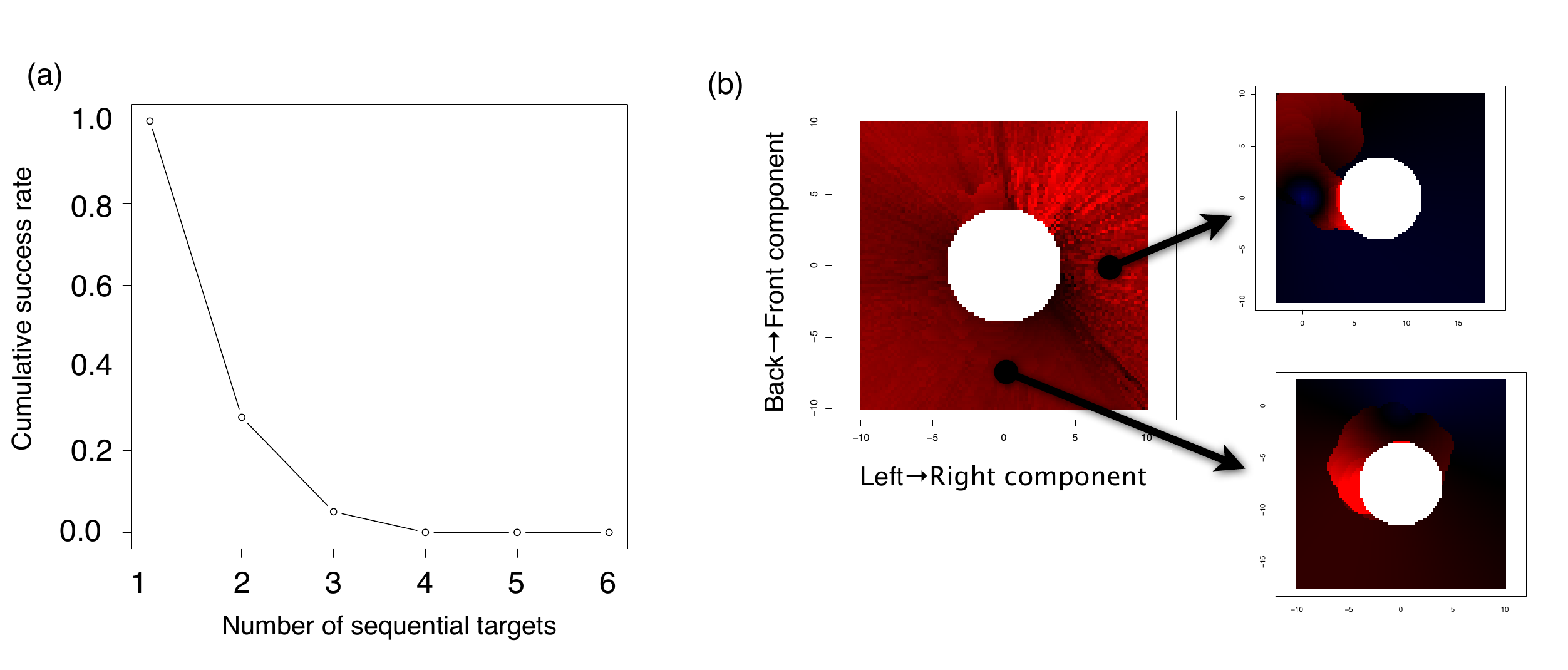}
			
		\end{center}
		\caption{
		{\bf First efficient one-food-source forager: key organism 6809.} 
		 (a): Despite its flawless ability to reach any food source on the grid once,
		this individual misses each following following food source  75\% of the time. 
		This is likely due to a switch in the organism's internal state after it reaches the first food source. (b): 
		Foraging map (left) and conditional foraging maps (right, conditional on two first-food-source locations reached). 
		The switch in internal state, exemplified by the drastically changed foraging map, 
		is moreover conditional on the previous food source position, as shown by the arrows.
		}
		\label{fig:foraging_6809}
		\end{figure*}

		Typically, the addition of noise to the target positions is an effective means to improve the forager's coverage. With little or no noise, 
		organisms  show a reduced ability to forage in their environment, being generally constrained to the close vicinity 
		of the four target positions they were exposed to. The target positions that are within the organism's reach
		are typically clustered together, forming one ore more patches as revealed by the foraging map. As the organism's foraging ability improves, the patches grow
		and eventually coalesce into bigger connected areas. 
		The remaining targets that are still out of reach form compact patches that the organism does not seem to be able to approach. 
		For food sources located in such areas, the organism remains stationary, although it actively wiggles using a gait that
		is superficially similar to the efficient one used to reach targets. 
		That inefficient gait, however, seems to be completely insensitive to the targets:
		the organism acts as if it could not sense their position (see Fig.~\ref{fig:confusing_zone}), as if the food sources were in a ``blind spot''. However, as the targets are always sensed regardless of their position, the term ``zone of confusion'' is more accurate. The relationship between the efficient and inefficient gaits will be discussed in more details in section~\ref{lab:distinctive_behaviors}.

The amount of noise used to evolve the first key organism was set to 0.1\% (see Figure \ref{fig:LOD}), which corresponds to a maximum perturbation of only $\pm$ 1 cm (as targets are positioned 10 meters away from the organism before noise is applied) along both planar coordinates (Fig.~\ref{fig:noise_application}). The amount of noise is drastically increased in each transfer to 5\% in the first transfer, and 50\% in the second, after which the positions are uniformly random. We found that if the noise is increased too fast, the overwhelming majority of the runs' fittest organisms wander about randomly. After the first transfer, when promising organisms are selected from the first-stage runs and are used to seed a new evolutionary run, their ability to cope with noise seems to increase exponentially. After two transfers, starting from a noise level of 0.1\%, the organisms are typically able to cope with noise up to 20\% to 50\%. Past those noise levels, they are able to reach food sources positioned uniformly around them, without missing any.  
		
After each evolutionary stage, the elites of each run improve their foraging ability in one of two qualitatively different ways: either they were able to reach more distant food sources, or they evolved the capacity to forage in a new direction. Another clear trend is that evolving the ability to forage in new directions is a rare occurrence compared to evolving the ability to exploit existing directions by reaching farther targets: turning appears harder to evolve than running. For this reason, we selected organisms primarily for their ability to reach food sources in all directions rather than reaching distant ones. 
		
Despite the advantage of allowing organisms to forage over greater areas, using noise carries an important drawback: it alters the organisms' fitness and rankings between generations. Not only is the best organism's fitness in one generation different from the fitness value in the next generation, but it might not be the fittest anymore. Since the food source positions are randomized for each new generation, the organism's fitness fluctuates throughout an evolutionary run. The fluctuations are more drastic as the noise level increases, sometimes over ten orders of magnitude (see for example generation 7 in Fig.~\ref{fig:organisms_fitness}d). This upward trend in fitness fluctuations is due to the enormous amount of fitness that efficient foragers are able to accumulate. Under this regime, it is hard to design a selection scheme that consistently favors promising foragers across generations. In previous work when evolving forward locomotion or throwing~\cite{ChaumontEgliAdami2007,ChaumontEgliAdami2006}, transferring the best organism along with others chosen with roulette selection to the next generation was sufficient to allow a monotonic increase in fitness within an evolutionary run. Almost always, higher fitness meant better, that is, more functional, organisms. In the evolution of autonomous foragers, this is not true any more: even at very low noise levels, selecting only one elite does not guarantee that good genes are going to be passed on to the next generation. Instead, as discussed earlier, the fittest organism could be the result of chance alone, where a food source happens to be placed directly on the individual's path. Chance can, however, be mitigated by increasing the number of elites. We found that five elites seem to be the threshold at which there is a marked difference in the algorithm's ability to find efficient organisms. We used eight elites in order to further increase the probability of selecting good genes, while the rest of the survivors were chosen with roulette selection. Under these conditions, the optimization algorithm manages to significantly increase the organism's foraging efficiency because the rankings within a population are then consistently preserved: A good forager at a given generation is also going to perform well in the next generation.

After two transfers, the best of the evolved organisms across the repeats is able to forage in all directions (see Fig.~\ref{fig:foraging_6809}b), and can reach any target on the test grid within a minute of simulated time (i.e., $60\cdot50=3000$ time steps). Interestingly, after reaching the first food source, the organism reaches the next target only a quarter of the time on average, and never reaches four or more (Fig.~\ref{fig:foraging_6809}a). Indeed, after reaching the first target, the organism's foraging map switches to a different one that has substantial zones of confusion (Figure \ref{fig:foraging_6809}b). To achieve autonomous and sustained foraging, an organism has to be able to collect food sources multiple times in sequence. As our goal is to use such a forager  as a seed ancestor to self-reproduce as it gathers energy from food, and ultimately seed a sustainable population in an open environment, we have further evolved the best organism to reliably forage several food sources in sequence.
		
	\subsection{Reaching more than one food source in sequence} 
		
		If organism 6809's foraging ability were not conditional on the food sources reached, it would forage \emph{all} of them,
		regardless of whether they are the first or the subsequent ones, and would produce a foraging profile in Fig.~\ref{fig:foraging_6809}a
		with a line close to 100\% success rate.
		However, its behavior as shown in Fig.~\ref{fig:foraging_6809}b is drastically altered upon reaching the first food source.
		The conditional foraging maps are qualitatively similar independently of the location of the first source, in that the organism attains on average only 25\% of the second food sources, as opposed to 100\% for the first foraging map. 
		
		This observation is important for the rest of the discussion: organism 6809 and its descendants exhibit a clear qualitative difference between
		the initial foraging map (which corresponds to the organism's ability to forage the first food source)
		and the conditional foraging maps that capture the organism's ability to reach the \emph{subsequent} food sources. 
		Each foraging map corresponds to a probability that an organism reaches a target (the success rate in Figure~\ref{fig:foraging_6809}a).
		For one food source, this probability is equal to the success rate (see Fig.~\ref{fig:foraging_6809}a) because the initial foraging map is unique. The success rate for reaching multiple food sources is contingent on reaching the previous ones, so it can never increase: the success rate to reach two food sources is the probability to reach the first multiplied by the probability to reach the second. 
		For each repeat in Fig.~\ref{fig:foraging_6809}a, the same organism reaches the first food source at different positions,
		so the conditional foraging map is \emph{different} each time. 
		The second probability (leaving the first food source to reach the second one) is an average of \emph{all} the probabilities associated 
		with all the conditional foraging maps.
		This second probability here is approximated by the ratio between the success rate for reaching the second food source divided by the success rate for the first one. The same logic applies to the success rate for three or more food sources. For example, in Fig.~\ref{fig:foraging_6809}a, the ratios between two consecutive success rates (from the first to the fourth food source) 
		yield low values of $0.28$, $0.179$, and $0$, respectively. These numbers mean that the organism is able to move on to the next food source only 28\% of the time after reaching the first food source, 17.9\% of the time after reaching the second one, and never after reaching the third. These percentages are clearly in stark contrast with the ability to systematically attain the first target.
		
		Had the experimental setup placed enough selection pressure on the ability to reach more than one food source,  the conditional foraging maps would have displayed more red pixels. In this section we describe our shift from the fitness function used in the first three stages to a different one 
that increases the evolutionary pressure on the organism's ability to forage more than one food source. Since our best individual after two transfers is able to reach the first food source systematically, we dropped the reward for it:  only the second and subsequent targets are rewarded.  
		This change corresponds to altering the fitness term in Eq.~\ref{eq:Wa} so that the outer product skips $W_s$ for $s=1$
		and factors the remaining fitness components for $s=2 \ldots S$. The fitness function $W$ thus becomes:
		\begin{equation}
		\label{eq:Wrepeat2}
			W = W_l \cdot \prod_{s=2}^{S}{W_{s}} + W_r
		\end{equation}
This alteration corresponds to the fitness function (b) in Fig.~\ref{fig:LOD}. At this stage, we also change the method to select the survivors: 
		we obtained better results for the new fitness function by switching from roulette to tournament selection with a tournament of size 5. 
		Finally, the food sources are now placed completely at random, even though there still are four evaluations per
		individual with a maximum of three attainable food sources per evaluation.
		
\begin{figure}[!ht]
		\begin{center}
			\includegraphics[width=0.45\textwidth]{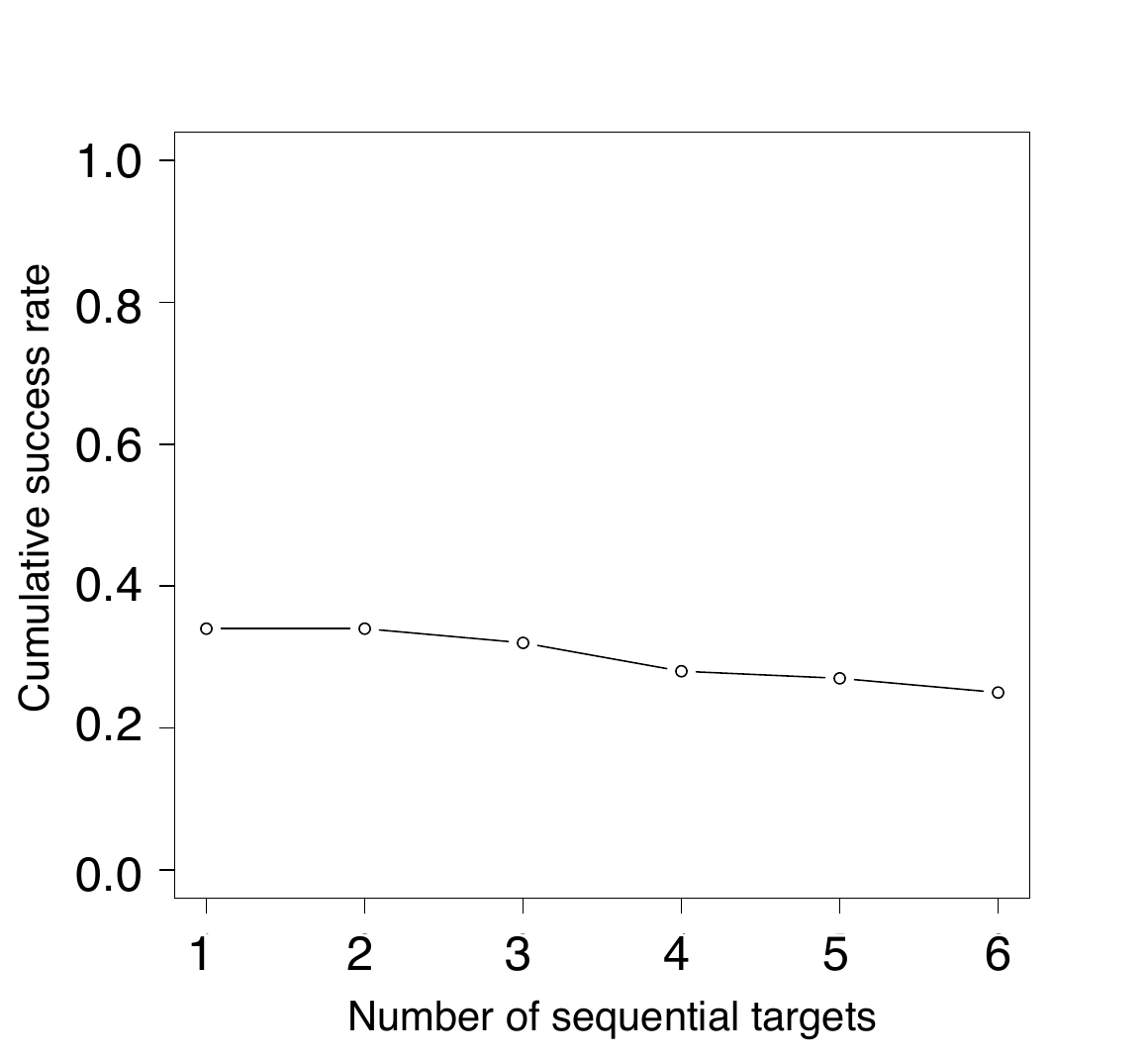}	
		\end{center}
		\caption{
		{\bf Foraging strategy that commonly evolved after rewarding the second food source onward, but not rewarding the first.}
		Although not reflecting a reliable foraging strategy, this profile is qualitatively the opposite 
		of its ancestor's in Fig.~(\ref{fig:foraging_6809}a): The first food source is not reached systematically
		anymore, but when it is, the following ones are attained very consistently. The morphology of this forager is shown in Fig.~\ref{fig:morph}, and its behavior on five consecutive food sources can be viewed  in a movie file that is part of the Supplementary Material.
		}
		\label{fig:foraging_profile_6981}
		\end{figure}

		Although a majority of the evolutionary runs in this stage led to organisms very similar to the ancestor (organism 6809), a few
		have a clearly different foraging ability (see, e.g., Fig.~\ref{fig:foraging_profile_6981}). This type of organism is able to reach the first food source only 40\% of the time, but once reached, it gets to the next one almost systematically. 
		The conditional foraging map for the second food source for this organism is qualitatively similar to that of the earlier ancestor's \emph{initial} foraging map (see Figure \ref{fig:foraging_6809}b). This suggests the hypothesis that different ``neural modules" are used by organism 6981 to reach the first as opposed to all subsequent food sources, and that the change in fitness landscape (removing a reward for the first food source) created organisms where the modules were simply swapped.  	

			\begin{figure}[!ht]
		
		\begin{center}
		\includegraphics[width=0.35\textwidth]{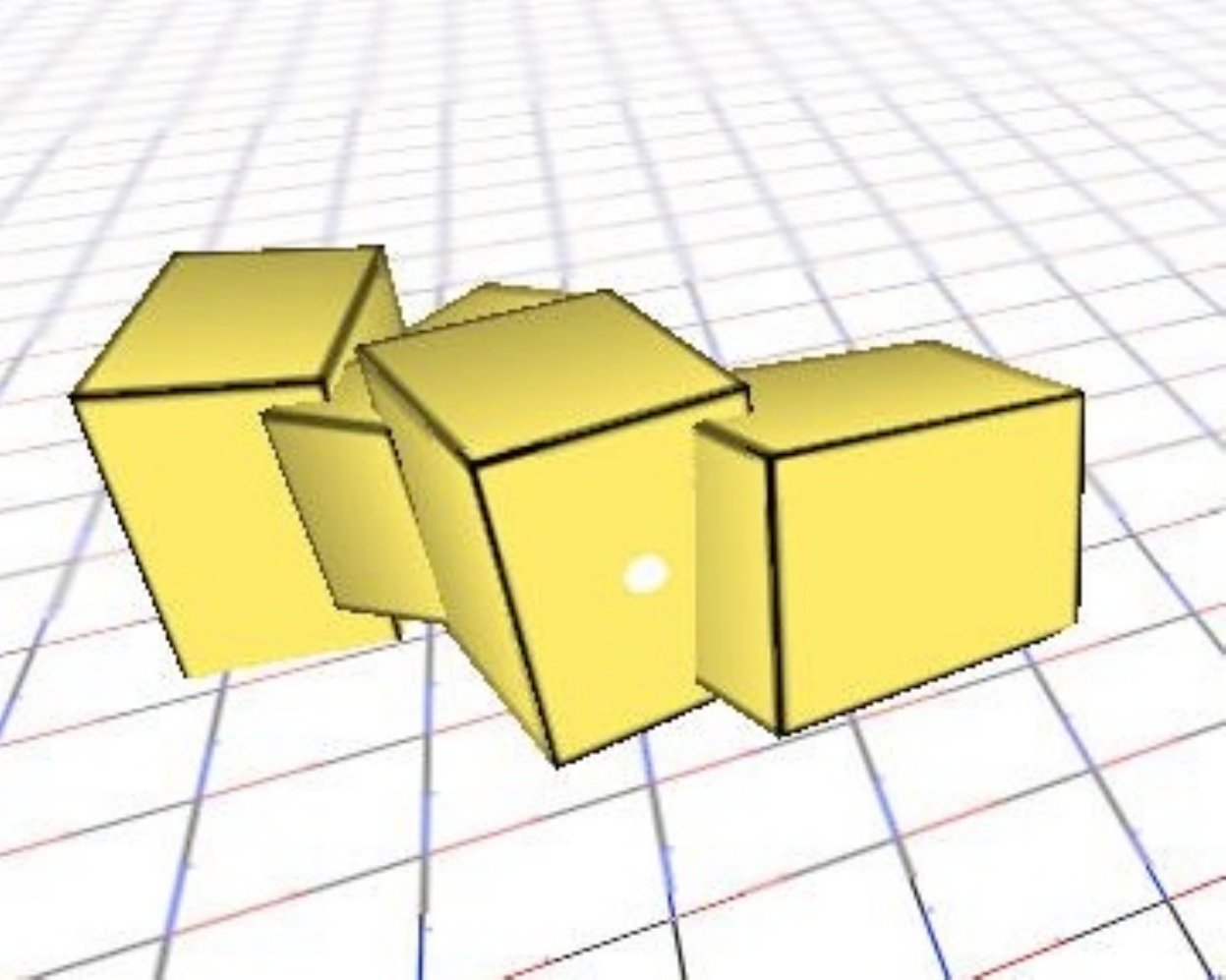}
			\end{center}
		\caption{{\bf Morphology of evolved forager 6981. }
		 Forager 6981 consist of 4 parallelepipedic blocks of varying dimensions, connected by 3 joints (indicated by whited dots). Adjacent blocks that share a joint are allowed to interpenetrate in order to allow for bending flexibility around the joint, but non-adjacent blocks (and adjacebt blocks not connected by a joint) cannot interpenetrate.}.
		\label{fig:morph}
				\end{figure}

		\begin{figure*}[!ht]
			\includegraphics[width=\textwidth]{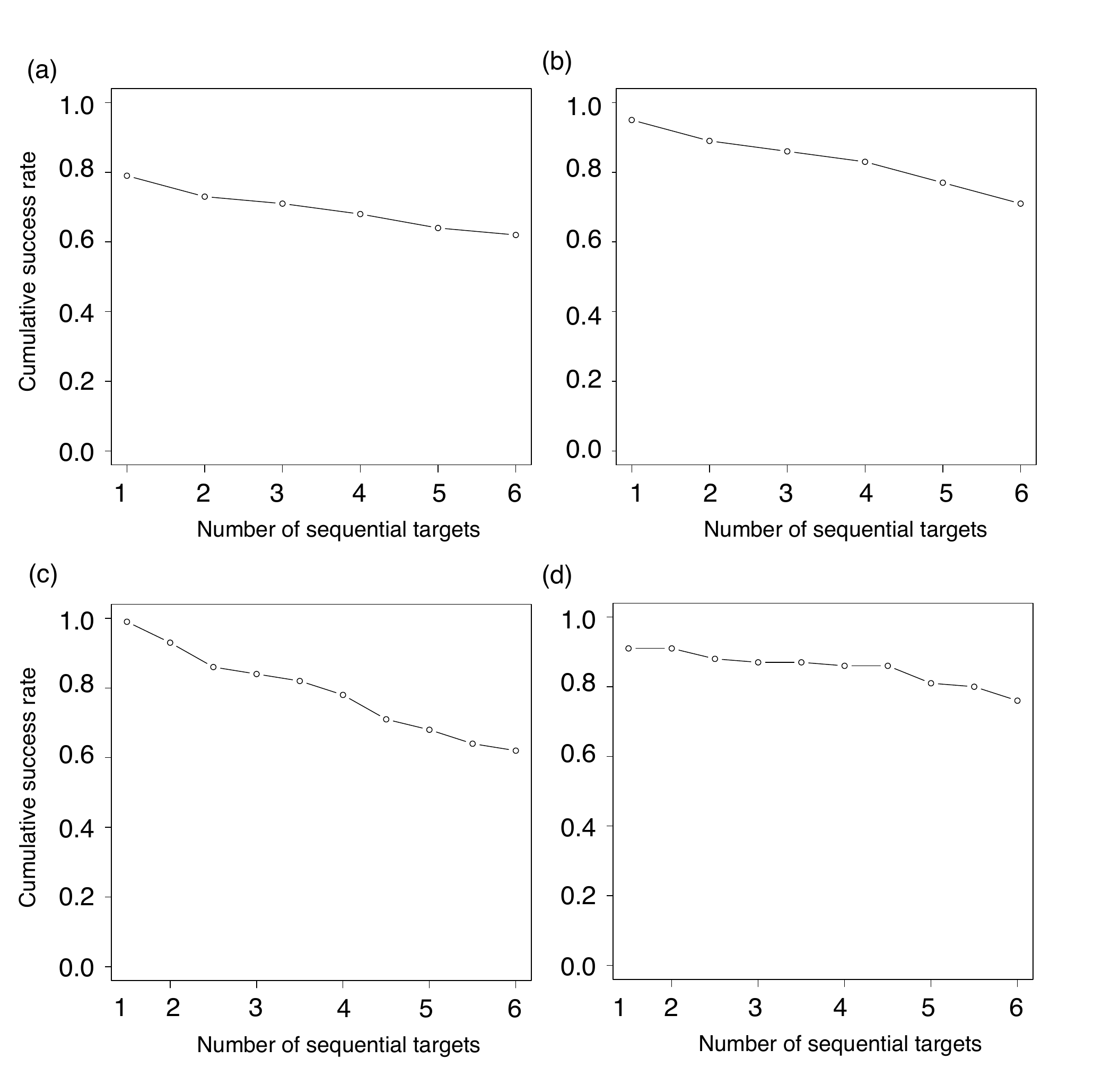}
		\begin{center}

		\end{center}
		\caption{
		{\bf Sequential foraging profile for the key organisms that are descendants of organism 6809.}
		Most of the runs seeded with organism 6809 yielded new strategies with foraging performances very similar to that of Fig.~\ref{fig:foraging_profile_6981}.
		(a): Organism 7046 is the only marked exception, with 79\% success rate at reaching the first food source. 
		(b): Its direct descendant, organism 7336  has a realized success rate of 95\%.
		The two last key organisms come from the same set of repeats, where organism 7336 was used as a seed.
		(c): Organism 7420 is the one that reaches the first food source most of the time (99\% success rate),
		whereas (d): organism 7513 has the lowest average miss rate (about 1.5\%).
		}
		\label{fig:foraging_second}
		\end{figure*}

		We observed one exception among the foragers that consistently forage subsequent food sources: 
		an individual (7046) was able to reach 80\% of the food sources initially instead of 40\%, and proceeds to the following ones 
		even more systematically (see Figure \ref{fig:foraging_second}a). Organism 7046 is used as the ancestor for the last transfer where the fitness function is simply the number of food sources reached across all the simulations (see Fig.~\ref{fig:LOD}). 
		The number of evaluations was also increased to six per individual for this last stage, with the same pattern of three sequential targets per evaluation (fitness function (c) in Fig.~\ref{fig:LOD}).
		The organisms evolved from this ancestor are significantly better foragers, and usually reach
		the first food source about 90\% of the time and miss very few afterwards. We identified two variations of very fit organisms: 
		The first type reaches the first food source the most often (99\% of the time, see Fig.~\ref{fig:foraging_second}c), but misses the next ones about 4\% of the time on average. 
		The second type (Fig.~\ref{fig:foraging_second}d) has the lowest miss rate for the following food sources of about 1.5\% on average,
		but reaches the first food source less often (91\% of the time). 		
		If evaluated for 10 food sources, the first and second types's total rates of success 
		are about 86\% and 85\% respectively. Note that we never selected for reaching more than three food sources in sequence, but efficient foragers reach an arbitrary number.

\section{Discussion}
\label{lab:distinctive_behaviors}
		Unlike undirected or directed locomotion, ground foraging proved to be much harder to evolve and posed several problems 
		that could not be anticipated from our previous experience with the systematic emergence of locomotive gaits~\cite{ChaumontEgliAdami2006}.
		The first difficulty is to evolve behaviors conditional on the food source positions, which requires to place the targets in opposite directions.
		One might argue that this behavior might not be so hard to acquire if the organisms were allowed to evolve for longer than only 40 generations,
		but in comparison, forward locomotion emerges within five generations. 
		In addition to saving CPU power, the small number of generations mitigates the problem of over-fitting the population to a particular fitness peak.
		It does not completely prevent over-fitting, however: efficient foragers in a two-food-source-setup such as illustrated in Fig.~\ref{fig:evaluation_steps}b, when transferred to an environment with four targets (Figure~\ref{fig:evaluation_steps}c) in a fashion similar to behavior chaining \cite{bongard2008behavior}, 
		are far less effective at going to the additional food sources than evolving organisms with the four food sources at the outset.
		The forward and backward wiggling behaviors, although well-suited for the two-food-source setup, appears to occupy a very narrow fitness peak: when mutated, the organism's fitness drops so much that the mutant cannot survive for long enough to be reconfigured into a potentially more effective forager able to turn right and left.
		This example illustrates that pre-evolved behaviors do not necessarily facilitate further evolution, 
		even though the skills acquired in the previous environment are readily useful and necessary for success in the new environment.
		
		Only very little noise could be applied to the food source positions in order to evolve directed foraging in the first 40 generations.
		As opposed to random wandering, directed foraging is optimal if the information provided by the food source positions helps the foragers to improve their
		fitness significantly: the noise in the food source placement is translated as random fitness fluctuations in the population, and the organism's attempts to approach food sources aims at reducing this fluctuation. If the noise results in perturbations that exceed the maximum distance by which 
		an organism can approach a food source, then the organism's foraging skill is useless since its fitness is dictated by the target positions.
		Selection cannot detect the organisms that exhibit intermediate foraging abilities, and this behavior is therefore not selected for. 
		As a consequence, the chances of seeing an efficient forager with complex neural wiring that emerges from scratch becomes effectively nil, 
		and evolution instead takes an easier path and improves the individual's body coordination to traverse greater distances in random directions, 
		without processing any information from the target positions.  
		
		Even when the noise is low enough that very few simulations yield randomly behaving organisms, the fraction of behaviors
		that visually look promising is very sensitive to the noise level.
		For example, in an initial set of 150 replicates evolved for 40 generations with 0.1\% of noise (not part of the set studied here), 5 organisms were clearly trying to move in at least three different directions, and were deemed to be worthy ancestors for the next set of repeats. 
		If the noise level is raised to 5\% with otherwise the same experimental setup, no interesting organism emerges in 200 replicates, and if the noise is raised to 20\%  and the number of replicates is increased to 500, no interesting behavior arises even after 100 generations. 
		We suspect that this extreme intolerance to noise in the early generations is a property peculiar to the type of neural controller used in this work.
		It would be interesting to map the evolution of a successful forager's controller resilience to noise across its line of descent, 
		and see how mutations reconfigure the neural network up to the point where it acquires tolerance to arbitrary noise.
				
		Once an efficient one-food-source forager is discovered, we expected it to be able to reach the next food source with the same efficiency. Instead, we found that foraging one food source,  and sustained, repetitive foraging, appear to be two entirely different skills. 
		Acquiring the former does not guarantee the emergence of the latter, as can be seen in organism 6809 (Fig.~\ref{fig:foraging_6809}).
		From the organism's foraging profile in Fig.~\ref{fig:foraging_6809}a, it is tempting to conclude that the organism operates in two distinctive modes: One that allows it to reach any food source initially, and another (where it misses 75\% of the time) that appears to be used for all the remaining food sources (explaining the power-law decay of foraging efficiency). The foraging profiles for the descendants 
		6981 (Fig.~\ref{fig:foraging_profile_6981}), 7046 (Fig.~\ref{fig:foraging_second}a),  
		7420 (Fig.~\ref{fig:foraging_second}c), and 7513 (Fig.~\ref{fig:foraging_second}d) also support this conclusion: 
		Regardless of the initial success rate at reaching the first food source, the organisms seem to miss the following ones at a similar rate.
		While looking at the foraging patterns for the first food sources for individuals 6981, 7046, 7420 and 7513, we notice a general trend 
		in the pattern for individual 6981 (Fig.~\ref{fig:Multi_foraging_maps_6981}) that looks like a red flare or smoke morphs into the intermediate pattern 
		of individual 7046 (Fig.~\ref{fig:Multi_foraging_maps_7046}) as if it was wrapped around the center, and slightly differentiates into two even 
		more aggressively wrapped versions in individuals 7420 (Fig.~\ref{fig:Multi_foraging_maps_7420}) and 7513 (Fig.~\ref{fig:Multi_foraging_maps_7513}).
		A similar observation applies to the conditional foraging maps for the second food sources: 
		across these four organisms, we can see that except for the map that is mostly black in Fig.~\ref{fig:Multi_foraging_maps_6981},
		all the maps share the same flare-like patterns that radiate away from the center in a straight direction, 
		as opposed to the whirling patterns described earlier.
		The descendants  from organism 6981 conceivably evolved two qualitatively distinct behaviors, one for reaching the first food source, 
		and another for the remaining food sources, and both apparently evolved separately. 
		
		A key behavior switch seem to have occurred between organisms 6809 and 6981: 
		the flare-pattern in one of organism 6809's \emph{conditional} foraging maps (upper right corner in Fig.~\ref{fig:foraging_6809}b) 
		surfaces in its descendant's \emph{initial} foraging map almost identical, only rotated. 
		Conversely, the flare-like pattern found in organism 6809's \emph{initial} foraging map seem to be expressed in two of its 
		descendant's \emph{conditional} foraging maps (upper left and lower right corners in Fig.~\ref{fig:Multi_foraging_maps_6981}).
		This observation together with the, apparently independent, modification of each behavior suggest that they might be 
		implemented by the neural network as two distinct functional modules that can be swapped and evolved individually, 
		which appears to be precisely what evolution took advantage of.
		
		The patterns described so far appear to be overlaid on simpler ones that seem to be independent of the target positions.
		For instance, the bottom right foraging map's background in Fig.~\ref{fig:foraging_6809}b shows a gradient pattern that is most probably explained by a single forward motion, regardless of the target position. 
		Such a constant motion might be a remnant from the ancestral gait that yields a basal fitness from forward motion alone [Equation~(\ref{eq:Wl})]. 
		This ancestral behavior remains because until the second-to-last transfer, we kept the term that rewards forward motion in the fitness function. 
		Blind forward motion has evolved first, and provided a building block upon which more complex behaviors have likely emerged. 
		We hypothesize that when a food source position does not trigger the efficient foraging behavior, the organism instead uses the (likely modified)
		ancestral forward gait as a fallback behavior, causing the individual to move invariably towards the same arbitrary direction.

\section{Conclusion}

		Despite its omnipresence in nature, the evolution in silico of efficient legged foraging that can be put on par with that of animals (such as small mammals) has remained out of reach.
		Inching towards the reproduction of such behavior in artificial media, we have presented here the first instance of efficient
		three-dimensional virtual-legged foragers that can reach food sources in a sustained fashion in their physical environment.
		The positioning pattern of the food sources, coupled with an adequate gradual increase of noise in their positions, 
		can drastically reduce the number of necessary targets without compromising the emergence of efficient foragers. 
		Those organisms have been shown to effectively use their sensory information to steer towards the food sources, 
		and they do so at increasingly arbitrary positions as they are further refined by evolution.
		
		The organisms' ability to move towards a food source can be visualized on a two-dimensional plot (a foraging map) that gives an instant 
		overview of the complex behavioral patterns displayed by foragers when tested on thousands of different food source positions.
		Those foraging maps help explain the foraging efficiency on successive food sources, and provide valuable insight into how the foraging
		behavior changes from ancestor to descendants. The conditional foraging map for the second food sources show how an organism's 
		ability to reach a food source at the same relative position can drastically change depending on the position of the first food source.
		Surprisingly, evolved foragers that are efficient at reaching the first food source do not perform consistently when attempting to reach multiple targets, and appear to switch their behavior after reaching the first food source. 
		Those organisms, when evolved further,  improve their overall efficiency and consistency, but conserve those two distinct behavioral modes.
		
		Despite drastic fitness fluctuations induced by imposing noise on food source locations (even between two consecutive generations) the Genetic Algorithm was able to
		detect significantly fitter organisms across generations so that evolution was able to improve the foraging performance substantially. This gave rise, after less than 300 generations of total evolution along the line of descent, to multiple foraging strategies that can reach the first food source in over 90\% of the cases and miss the following ones at most 4\% of the time on average.
		In future work, it would be instructive to analyze the forager's neural networks to study how they are modified on a per-generation
		basis, and compare those changes to how the foraging map is altered. 
		Such an analysis might provide insights to the origin and nature of the behavior switch displayed by organism 6809 and its descendants.
		However, such an analysis would be computationally expensive: the evaluation of the foraging maps alone across approximately 300 generations would take about two CPU-months for a resolution of $101\times101$ food sources. 
		
		It is possible that much of the dynamics that we report here are specific to the neural controller used, a controller that is composed of very high-level functions packaged into a single neuron. The evolvability of such a neural network can be questioned, as usually only a few neurons actually participate
		in a complex behavior (most neurons are neutral to knockout and therefore appear to be without function). The few functional neurons offer few and often very brittle degrees of freedom for evolution to act on. It would also be instructive to use exactly the same tools and the same parameters to evolve foragers with different types of controllers, such as continuous
		time recurrent neural networks, as in Refs.~\cite{Beer1996,Beer2003,bongard2008behavior}, or radically different ones 
		such as the Markov networks in~\cite{Edlundetal2011} and see what previous evolutionary obstacles they ease, or which new ones they impose.
		
		The most immediate usage of these robust evolved foragers is to simulate a population of clones in an open-ended environment, 
		to see if by gathering enough food items, individuals can reproduce and give rise to a sustainable population. 
		Such a confirmation would pave a straight path to ecological experiments
		where adaptive radiation can give rise to diversified ecotypes, and ultimately distinct species, among other experiments. 
		The promise for evolutionary experimentation of such a population is tremendous. 
		For example, the morphology or behavior of one type of organisms might provide substantial constraints that could prevent inter-species gene flow
		via hybrid inviability with other organism's morphologies or behaviors. 
		Such Dobzhansky-Muller incompatibilities (see, e.g.,~\cite{OrrTurelli2001}) have never been based on evolved morphologies or behaviors.
		Mate selection could also be grounded in physical realism: organisms may conceivably choose their mates
		based on morphological clues only, that could eventually be combined into species-specific signatures that reflect a specialized physical 
		apparatus that evolved to allow its bearer to survive in a specific niche~\cite{BolnickFitzpatrick2007}. 
		In fact, studying the co-evolution of morphology and behavior, and its impact on the dynamics of speciation, 
		is probably the most tantalizing prospect for experiments with the open-ended evolution of these creatures.
		Geographic separation, as well as various landscapes with different available resources could induce
		groups of organisms to geographically split, but also evolve the ability to take advantage of distinctive types of landscapes through 
		the development of appropriate physical and behavioral features. 
		
		The emergence of stable ecosystems would allow us to study questions in ecology, such as food web dynamics with the added 
		degree of physical and behavioral realism. Metabolic rate, behavioral, organism size, and animal range size variations to name a few,
		would all be naturally simulated for free. Previously, all those variables have only been modeled as mathematical abstractions. 
		However, even if several tens of species emerge, the total number of individuals needed to keep the ecosystem stable has to be substantial,
		numbering in the order of a thousand at least. Even though the current version of the software is able to simulate about a hundred of organisms
		on a single core, the physics engine needs to be more optimally used, and EVO needs to be parallelized to scale to one or two more orders
		of magnitudes and simulate an ecosystem on a single multi-core computer.

\appendix[Foraging maps for the first and second food sources for several individuals]
		\begin{figure}[!ht]
		\begin{center}
			\includegraphics[width=0.5\textwidth]{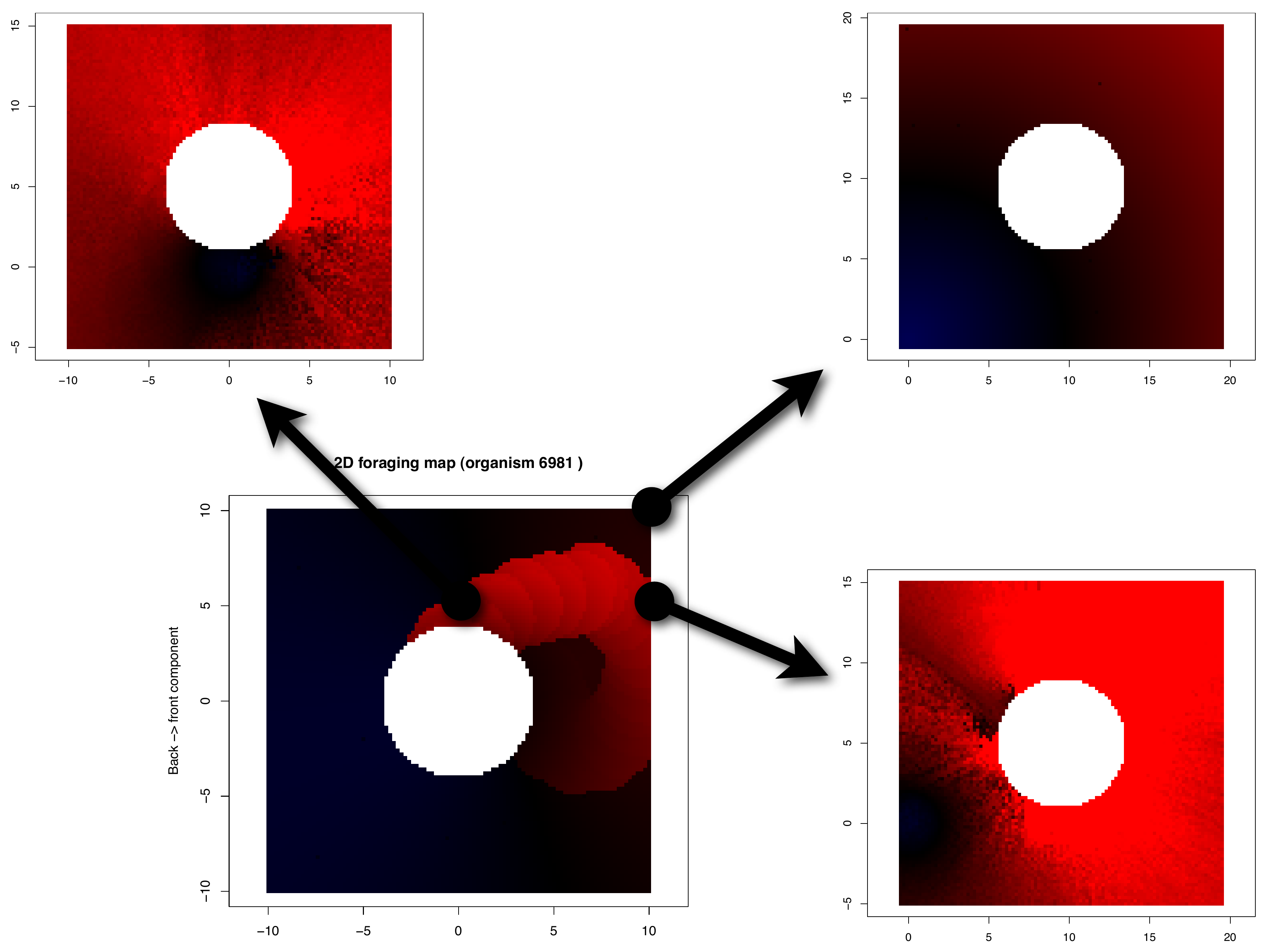}
		\end{center}
		\caption{
		{\bf Foraging maps for the first and second food sources for organism 6981.} 
		}
		\label{fig:Multi_foraging_maps_6981}
		\end{figure}

		\begin{figure}[!ht]
		\begin{center}
			\includegraphics[width=0.5\textwidth]{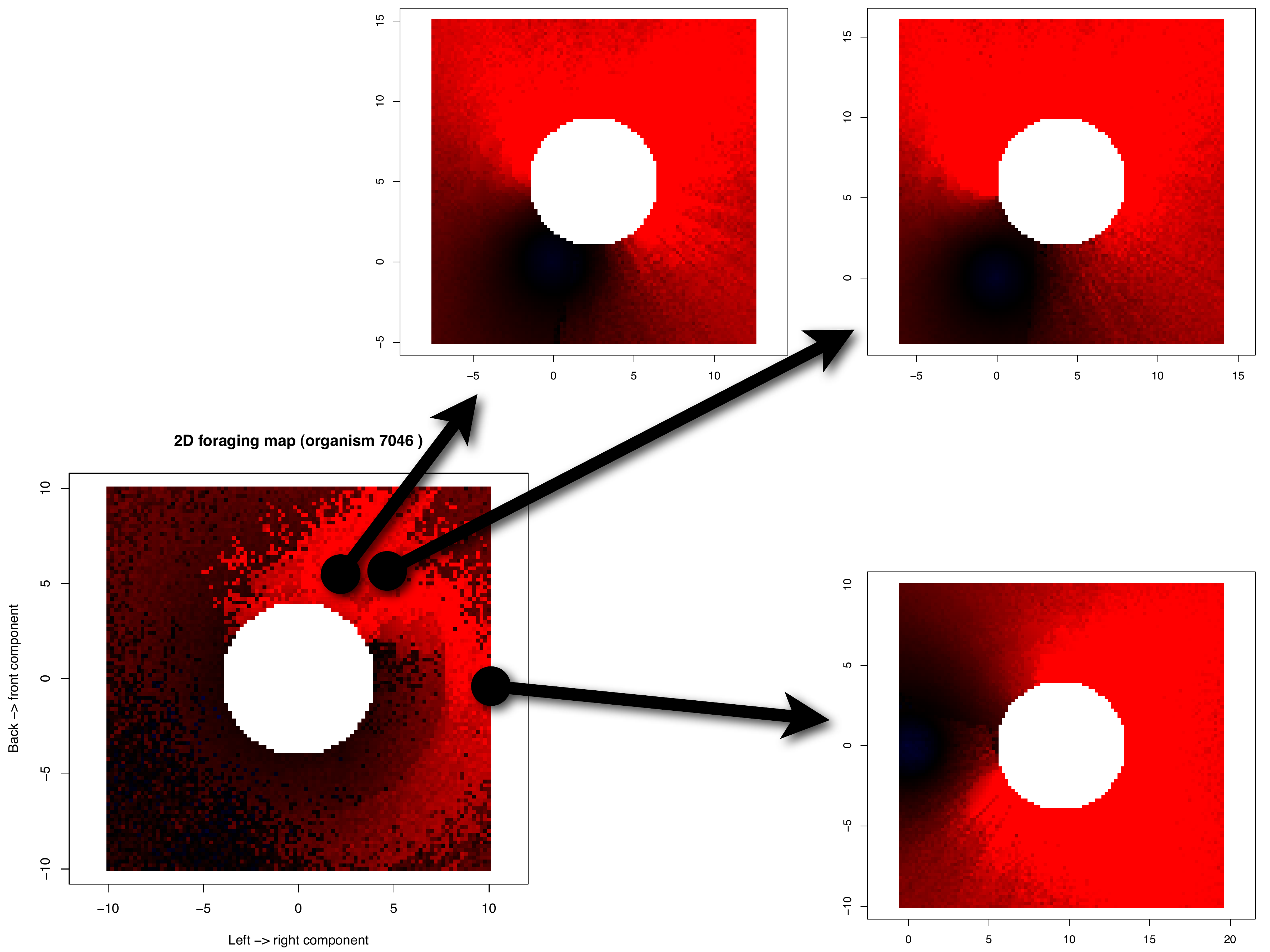}
		\end{center}
		\caption{
		{\bf Foraging maps for the first and second food sources for organism 7046.} 
		}
		\label{fig:Multi_foraging_maps_7046}
		\end{figure}

		\begin{figure}[!ht]
		\begin{center}
			\includegraphics[width=0.5\textwidth]{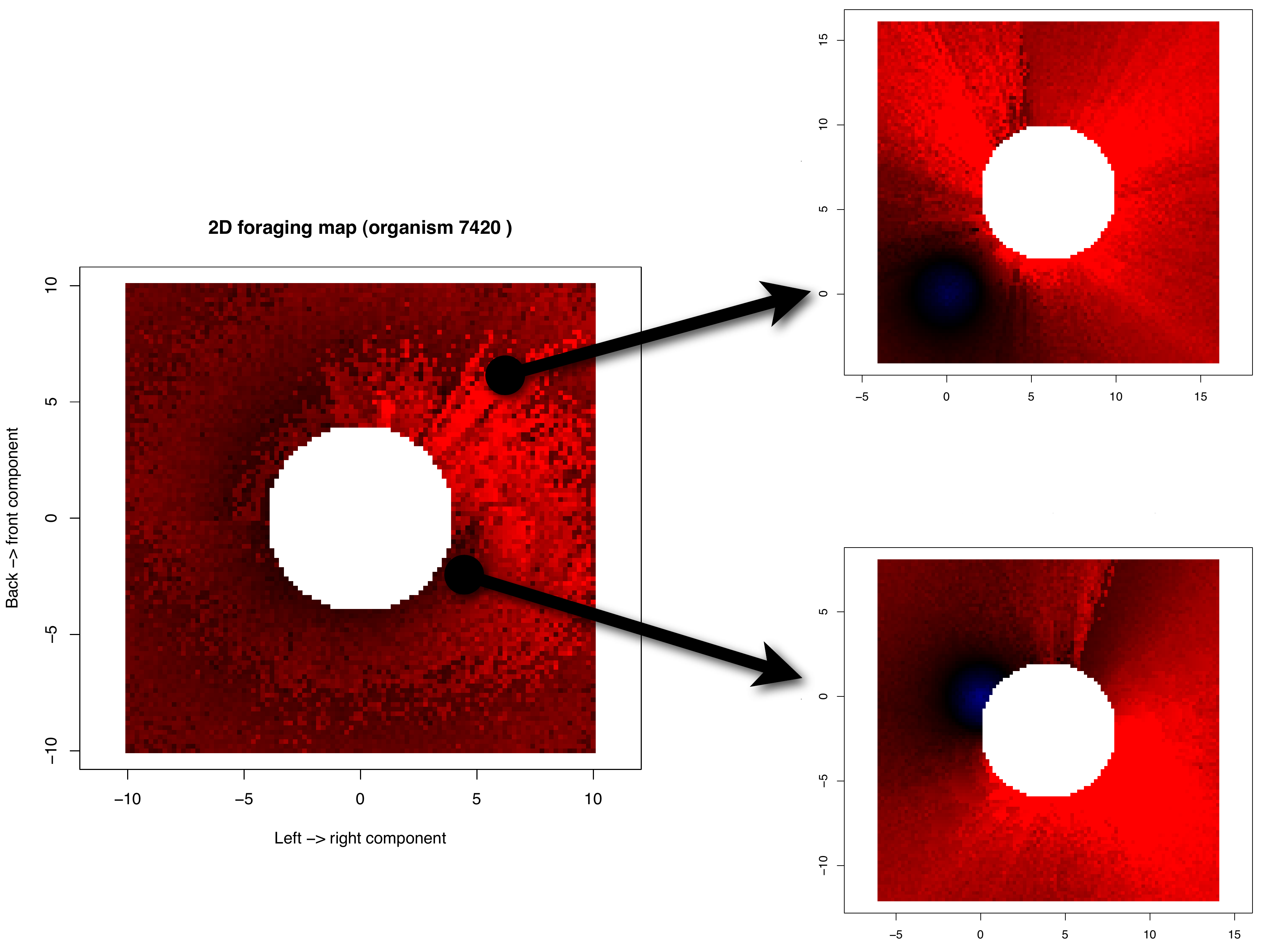}
		\end{center}
		\caption{
		{\bf Foraging maps for the first and second food sources for organism 7420.} 
		}
		\label{fig:Multi_foraging_maps_7420}
		\end{figure}

		\begin{figure}[!ht]
		\begin{center}
			\includegraphics[width=0.5\textwidth]{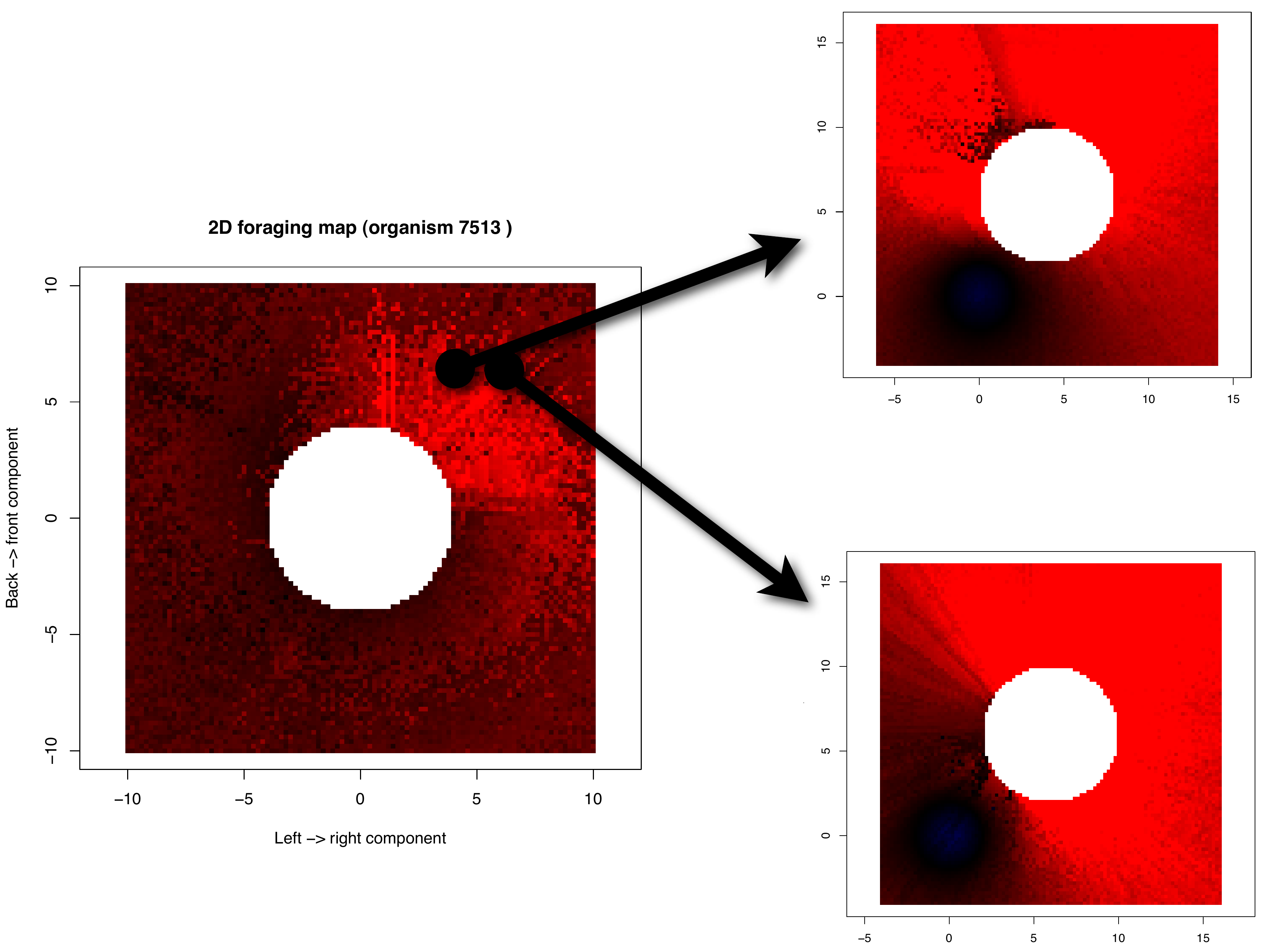}
		\end{center}
		\caption{
		{\bf Foraging maps for the first and second food sources for organism 7513.} 
		}
		\label{fig:Multi_foraging_maps_7513}
		\end{figure}

\section*{Acknowledgments}
	This research was funded in part by the National Science FoundationÕs Frontiers in Integrative Biological Research 
	grant FIBR-0527023, and the NSF BEACON Center for the Study of Evolution in Action under contract No. DBI-0939454. 
	The authors would like to thank the Keck Graduate Institute for the use of their computational infrastructure
	as well as Michigan State University's Institute of Cyber-Enabled Research for using their High Performance Computer cluster.

\ifCLASSOPTIONcaptionsoff
  \newpage
\fi



\bibliographystyle{IEEEtran}
\bibliography{IEEEabrv,Nicolas_Chaumont,Foraging_CA}
%



%

\begin{IEEEbiography}[{\includegraphics[width=1in,height=1.25in,clip,keepaspectratio]{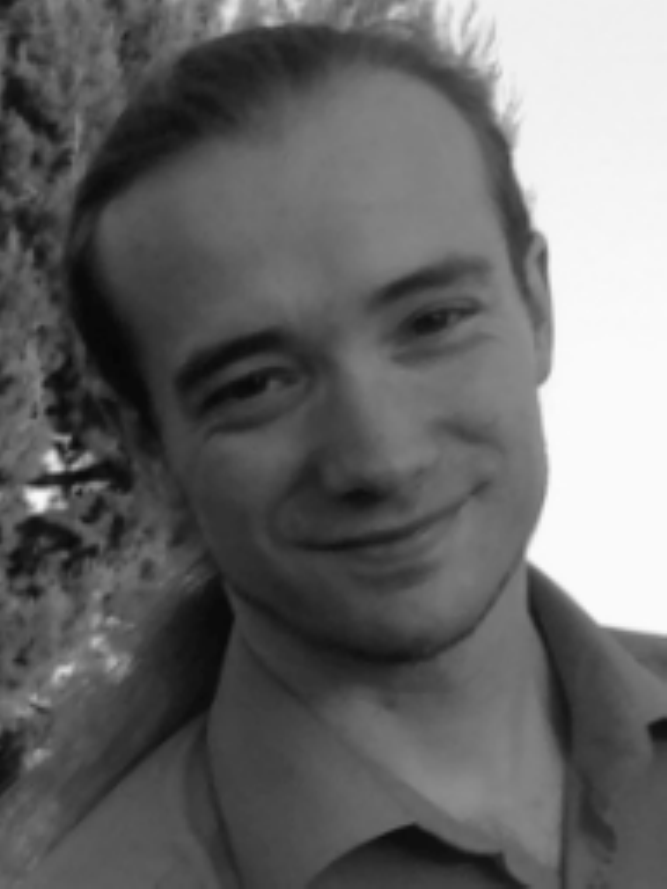}}]{Nicolas Chaumont}
received an M.S. in Computer Science from the University of Sherbrooke (Canada), as well as an M.A. in Mathematics from Claremont Graduate University and a Ph.D. in Computational Systems Biology from the Keck Graduate Institute (Claremont, CA). His current research interest include the co-evolution of morphology and behavior in animals, and how these traits affect the formation and stability of species and ecosystems. 
\end{IEEEbiography}

\begin{IEEEbiography}[{\includegraphics[width=1in,height=1.25in,clip,keepaspectratio]{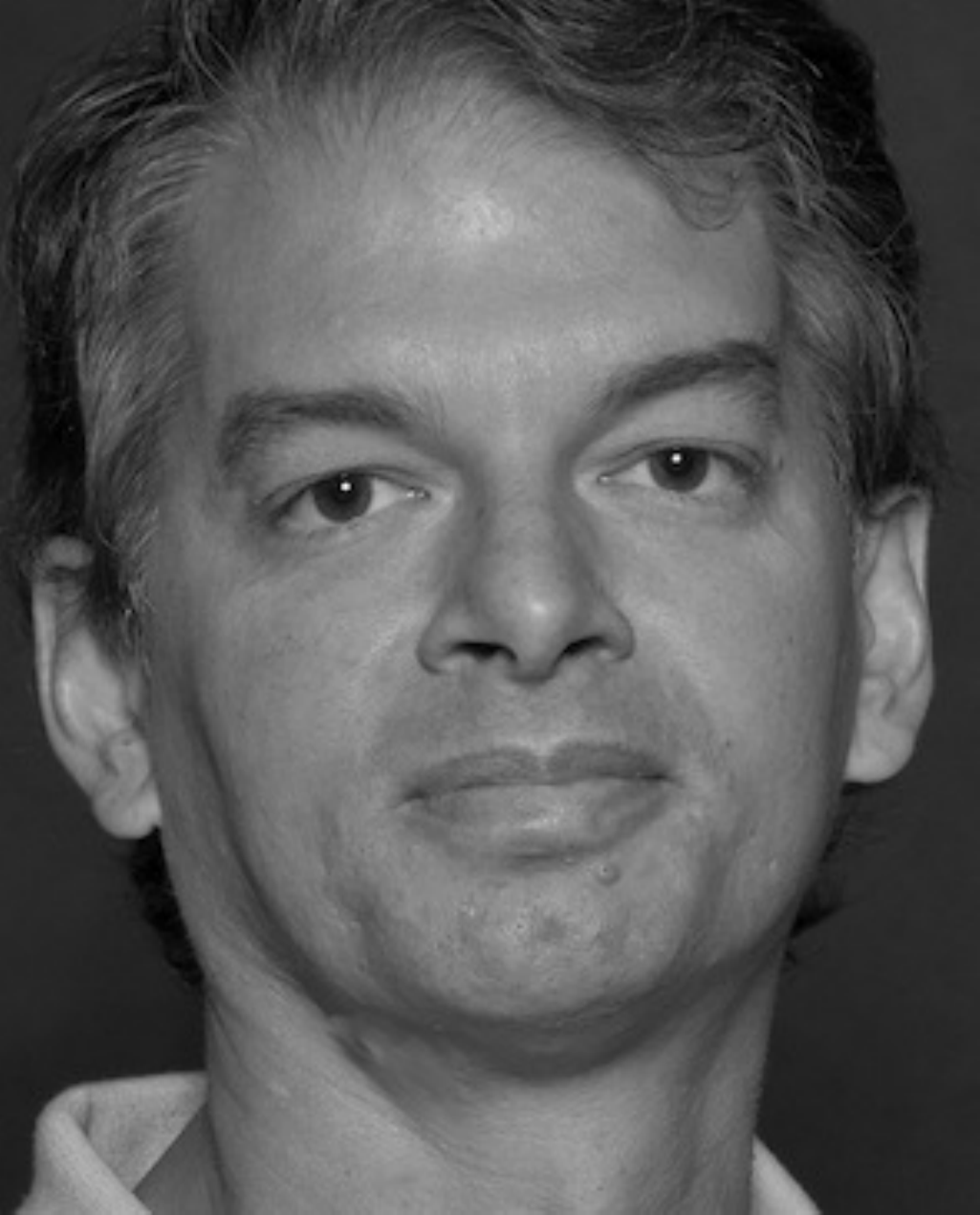}}]{Christoph Adami}
is Professor of Microbiology and Molecular Genetics, as well as Professor of Physics and Astronomy, at Michigan State University. He obtained his Ph.D. and M.A. in theoretical physics from the State University of New York at Stony Brook, as well as a Diplom in Physics from Bonn University (Germany). His main research focus is Darwinian evolution, which he studies at different levels of organization (from simple molecules to brains). He has pioneered the application of methods from information theory to the study of evolution, and designed the ÒAvidaÓ system that launched the use of digital life as a tool for investigating basic questions in evolutionary biology. He wrote the textbook ÒIntroduction to Artificial LifeÓ (Springer, 1998), is a fellow of the AAAS and the recipient of NASAÕs Exceptional Achievement Medal. 
\end{IEEEbiography}





\end{document}